# AGTCNet: A Graph-Temporal Approach for Principled Motor Imagery EEG Classification

**GALVIN BRICE S. LIM**[1] **(Student Member, IEEE), BRIAN GODWIN S. LIM**[2],
**ARGEL A. BANDALA**[1] **(Member, IEEE), JOHN ANTHONY C. JOSE**[1]**, (Member, IEEE),**
**TIMOTHY SCOTT C. CHU**[3]**, AND EDWIN SYBINGCO**[1] **(Senior Member, IEEE)**
[1]Department of Electronics and Computer Engineering, De La Salle University - Manila, Metro Manila 1004, Philippines
[2]Division of Information Science, Nara Institute of Science and Technology, Nara 630-0192, Japan
[3]Department of Mechanical Engineering, De La Salle University - Manila, Metro Manila 1004, Philippines

Corresponding author: Edwin Sybingco (e-mail: edwin.sybingco@dlsu.edu.ph).

This work was supported in part by the Department of Science and Technology (DOST) of the Philippines through the Engineering Research and Development for Technology (ERDT) program and in part by the Nara Institute of Science and Technology.

**ABSTRACT** Brain-computer interface (BCI) technology utilizing electroencephalography (EEG) marks a transformative innovation, empowering motor-impaired individuals to engage with their environment on equal footing. Despite its promising potential, developing subject-invariant and session-invariant BCI systems remains a significant challenge due to the inherent complexity and variability of neural activity across individuals and over time, compounded by EEG hardware constraints. While prior studies have sought to develop robust BCI systems, existing approaches remain ineffective in capturing the intricate spatiotemporal dependencies within multichannel EEG signals. This study addresses this gap by introducing the attentive graph-temporal convolutional network (AGTCNet), a novel graph-temporal model for motor imagery EEG (MI-EEG) classification. Specifically, AGTCNet leverages the topographic configuration of EEG electrodes as an inductive bias and integrates graph convolutional attention network (GCAT) to jointly learn expressive spatiotemporal EEG representations. The proposed model significantly outperformed existing MI-EEG classifiers, achieving state-of-the-art performance while utilizing a compact architecture, underscoring its effectiveness and practicality for BCI deployment. With a 49.87% reduction in model size, 64.65% faster inference time, and shorter input EEG signal, AGTCNet achieved a moving average accuracy of 66.82% for subject-independent classification on the BCI Competition IV Dataset 2a, which further improved to 82.88% when fine-tuned for subject-specific classification. On the EEG Motor Movement/Imagery Dataset, AGTCNet achieved moving average accuracies of 64.14% and 85.22% for 4-class and 2-class subject-independent classifications, respectively, with further improvements to 72.13% and 90.54% for subject-specific classifications. As guidance for future research, this study formally established practical model training-evaluation frameworks for subject-independent and subject-specific EEG classifications.

## I. INTRODUCTION

Brain-computer interface (BCI) marks a groundbreaking technological innovation, redefining human-computer interaction that transcends conventional limitations. This revolutionary technology directly converts brain signals into control commands, enabling the seamless transmission of user intentions to external devices [1], [2]. Among the various neuroimaging techniques, electroencephalography (EEG) plays a pivotal role in the realization of BCI systems owing to its noninvasive nature, ease of use, and portability. EEG measures electrical activity of the brain through electrodes positioned on the scalp, capturing neural patterns across different brain regions [3]. These patterns form the foundation of various BCI paradigms, wherein specific cognitive functions are associated with distinct brain regions and rhythms. In particular, this study focuses on the motor imagery (MI) paradigm [4], which involves mental simulation of motor actions without actual physical movements. This paradigm leverages the event-related (de)synchronization (ERD/ERS) patterns of neural signals, particularly within the alpha (8 Hz – 13 Hz) and beta (14 Hz – 30 Hz) frequency bands, which are regulated by the motor cortex [5].

In recent years, several studies have explored the integration of EEG-based BCI systems into assistive devices, such as wheelchairs [6], [7], [8] and prosthetic arms [9], [10], providing vital support for individuals with paralysis and physical impairments. Additionally, these systems have been incorporated into applications, such as drone control [11], smart home systems [12], [13], and entertainment platforms, including virtual and augmented reality [14], introducing new user experiences. This innovative technology not only offers unparalleled convenience but also serves as a transformative assistive tool, empowering individuals—particularly those with motor impairments—to engage with their environment on equal footing.

Despite their promising potential, EEG-based BCI systems remain in the early stages of development, facing numerous technical and practical challenges that hinder the realization of their full capabilities. A key challenge is the development of subject-invariant (subject-independent) and session-invariant (subject-specific) BCI systems capable of accurately classifying brain stimuli across diverse users and time periods. This challenge is primarily attributed to the intrinsic complexity of neural activity, characterized by billions of interconnected neurons that generate dynamic patterns, which vary across individuals and change over time depending on one's present mental state, emotional condition, and external influences [15]. Consequently, current BCI systems require user calibration, limiting their accessibility to the general public.

Moreover, the noninvasive nature of EEG also renders these BCI systems susceptible to various types of noise and artifacts, including involuntary stimuli and electrical interference, which distort neural signals. The limited number of electrodes and sampling rate of EEG headsets further constrain the spatial and temporal resolutions of neural signal measurements. These limitations pose significant risks depending on the application, as inaccurate signal classification could lead to unintended responses with potentially serious consequences. Hence, it is imperative to leverage information across all dimensions of multichannel EEG signals to enhance the BCI system's accuracy and mitigate potential risks.

While prior studies have developed various motor imagery EEG (MI-EEG) classification models, the majority rely on convolutional neural networks (CNNs) or recurrent neural networks (RNNs) to learn representations of EEG signals. These architectures overlook the non-Euclidean topology of multichannel EEG data, particularly the spatial correlations among adjacent EEG electrodes and the functional interactions within the brain, thereby failing to effectively leverage the rich spatial dependencies inherent in EEG signals. Recent advancements have sought to address this limitation by modeling EEG channels as a graph and leveraging graph neural networks (GNNs) to facilitate spatial feature learning. Nevertheless, existing graph-based spatiotemporal models generally employ a decoupled spatial and temporal feature learning architecture, which limits their ability to expressively capture the intricate spatiotemporal dependencies within EEG signals.

Thus, recognizing the inherent spatiotemporal nature of multichannel EEG signals, this study introduces the attentive graph-temporal convolutional network (AGTCNet), a novel graph-temporal model that jointly captures the spatial and temporal dynamics of EEG signals. By effectively learning subject-invariant and session-invariant EEG representations, AGTCNet enhances the generalizability of MI-EEG classification across diverse users and time periods.

The key contributions of this study are as follows:
- It introduces a novel graph-temporal architecture for subject-independent and subject-specific MI-EEG classifications, incorporating inductive bias through an EEG channel adjacency graph based on the anteroposterior and mediolateral topographic configuration of EEG electrodes to facilitate effective spatiotemporal feature learning from multichannel EEG signals.
- It develops the graph convolutional attention network (GCAT), which jointly captures the spatial and temporal features in sequential graph-structured data, thereby enabling the learning of robust and expressive spatiotemporal representations.
- The proposed AGTCNet model achieves state-of-the-art performance on both the BCI Competition IV Dataset 2a (BCICIV2A) and the PhysioNet EEG Motor Movement/Imagery Dataset (EEGMMIDB).
- It formally establishes practical model training-evaluation frameworks that account for subject-invariant and session-invariant considerations in real-world deployment scenarios, effectively narrowing the gap between offline and real-time performance and facilitating comparability between studies.

The remainder of this paper is structured as follows: Section II reviews related works. Section III introduces the proposed AGTCNet architecture, and Section IV establishes the practical model training-evaluation frameworks. Section V describes the evaluation methodology, followed by the presentation and discussion of the results in Section VI. Finally, Section VII concludes the study with a summary and recommendations for future research.

## II. RELATED WORKS

### A. EEG CLASSIFICATION MODELS

Over the years, BCI researchers have sought to enhance the decoding of MI-EEG signals through the application of traditional machine learning techniques and advanced deep learning approaches, with particular focus on the optimization of feature learning. These methods have been instrumental in learning the underlying information embedded within EEG signals, contributing to the advancements of BCI systems.

#### 1) TRADITIONAL MACHINE LEARNING MODELS

In EEG analysis, feature extraction methods, such as discrete wavelet transform (DWT) [16] and power spectral density (PSD) [17], are commonly employed to extract the temporal and spectral characteristics of EEG signals, transforming them into more expressive representations. Additionally, techniques like independent component analysis (ICA) [18], [19] and empirical mode decomposition (EMD) [20] are also used to decompose EEG signals into independent sources or intrinsic mode functions, effectively isolating the relevant components and eliminating artifacts, thereby enhancing the robustness of the extracted features. In particular, the common spatial pattern (CSP) algorithm [21], [22] has been widely recognized as a state-of-the-art method for MI-EEG feature extraction, owing to its ability to maximize interclass variance through spatial filtering. Building on this foundational approach, extended versions such as subband CSP (SBCSP) [23] and filter bank CSP (FBCSP) [24] have been developed, integrating spectral features to further improve classification performance. Furthermore, a separate study [25] applied EMD as a preprocessing step to CSP, further enhancing the robustness of extracted EEG features.

With these extracted features, supervised classification algorithms, such as linear discriminant analysis (LDA) [17], [20], [23] and support vector machines (SVM) [16], [19], [24], along with advanced variants like firefly-SVM (FASVM) [26] and regularized LDA (RLDA) [27], are extensively used for MI-EEG classification. While these algorithms, particularly when combined with CSP, have demonstrated competitive performance, traditional machine learning approaches heavily rely on feature extraction engineering, which limits their ability to model complex and high-dimensional data.

2) DEEP LEARNING MODELS

Recently, end-to-end deep learning models have gained popularity in EEG classification due to their ability to automatically learn underlying representations and achieve superior performance, establishing them as the prevailing state-of-the-art approach in the field. In general, 31.71% of the MI-EEG studies have utilized raw EEG signals with minimal preprocessing, such as standardization [28], [29], [30], [31], [32], [33], exponential moving average standardization [34], common average referencing [31], [32], bandpass filtering in the range of 8 Hz – 35 Hz [28] or 4 Hz – 40 Hz [33], [34], and downsampling [34], for classification [35].

Among deep learning models, convolutional neural networks (CNNs) have emerged as the most widely used architecture, with 78% of the studies utilizing this approach [36]. Both shallow [37], [38] and deep [38] CNN models have been employed, along with specialized variants like inception-based CNNs [28], [39] and multibranch CNNs [30]. Notable CNN-based models for MI-EEG classification include EEGNet [40], Deep ConvNet [38], Shallow ConvNet [38], EEGNet Fusion [41], and EEG-Inception [39], with EEGNet serving as a widely adopted benchmark in numerous studies. In particular, the EEGNet developed by Lawhern et al. [40] strategically employs temporal and spatial convolutions to capture features from multichannel EEG signals across both temporal and spatial dimensions.

In addition to CNNs, recurrent neural networks (RNNs), specialized for time-series data, have also been extensively used to capture the temporal dependencies within EEG signals. However, due to the limitations of standard RNNs, studies have utilized advanced variants such as multiplicative RNN (MRNN) [42], long short-term memory (LSTM) [43], and bidirectional LSTM (BiLSTM) [44], which offer better retention of long-term temporal dependencies and enhanced sequential modeling. For instance, Amin et al. [28] proposed a hybrid model that integrates inception-based CNN and BiLSTM to facilitate spatial and temporal feature learning. Moreover, temporal convolutional networks (TCNs), a CNN-based architecture designed for time-series modeling, have been adapted for MI-EEG classification due to their ability to expand the receptive field with minimal parameter overhead. Notable adaptations include EEG-TCNet [45], which integrates TCN with EEGNet, and TCNet-Fusion [46], further enhancing the performance through feature fusion techniques.

Inspired by selective attention in the human brain, recent advancements in MI-EEG classification have incorporated attention mechanisms to enhance classification performance by focusing on key features of the EEG signal. For instance, Altaheri et al. [29] enhanced EEG-TCNet and TCNet-Fusion by incorporating multihead attention (MHA) and ensemble learning, introducing the attention-based temporal convolutional network (ATCNet). Zhou et al. [30] later advanced this approach by employing a dual-branch architecture in the dual-branch attention temporal convolutional network (DB-ATCNet), achieving state-of-the-art performance. Additionally, self-attention mechanisms have also been incorporated into hybrid models, such as the convolutional recurrent attention model (CRAM) [47] and the global adaptive transformer [33], which combines CNN, LSTM, MHA, and generative adversarial network (GAN). Furthermore, owing to their ability to capture long-range temporal dependencies through self-attention mechanisms and efficiently process data in parallel, Transformers have been integrated into various MI-EEG models as well. For instance, EEG-Transformer [48], EEG Conformer [49], and ConTraNet [50] leverage Transformer encoders in conjunction with CSP, EEGNet, and CNN, respectively, to dynamically focus on the salient temporal segments of the EEG signal.

Moreover, to address the intersubject variability in EEG signals, models for subject-invariant feature learning have been developed. One such approach, as presented in [51], involves baseline correction, wherein linear layers are used to capture subject-specific features from the baseline EEG signals, which are then subtracted from the task-related signals to yield subject-invariant representations. Other studies have utilized autoencoders (AEs) to transform EEG signals into latent representations that capture subject-invariant features. Specifically, convolutional AEs were utilized in [52], whereas an LSTM-based AE was proposed in [53], accounting for the temporal dependencies inherent in EEG signals.

Furthermore, given the limited size of MI-EEG datasets, transfer learning has been adopted in previous studies [37], [54] to leverage EEG data from multiple subjects. This strategy facilitates the development of robust subject-specific EEG classifiers by adapting the learned neural dynamics from a diverse set of subjects to the nuanced characteristics of a particular subject. Despite these advancements, CNN- and RNN-based models remain ineffective in capturing the rich spatial information inherent in multichannel EEG signals.

## B. SPATIOTEMPORAL MODELS

Recognizing the spatiotemporal nature of EEG signals, recent studies have increasingly focused on developing spatiotemporal models to effectively learn EEG representations across both spatial and temporal dimensions. These models can generally be classified into pseudo-spatiotemporal and graph-based spatiotemporal models.

### 1) PSEUDO-SPATIOTEMPORAL MODELS

Pseudo-spatiotemporal models employ handcrafted techniques to preserve, to some extent, spatial dependencies within multichannel EEG signals. As demonstrated in [55], a straightforward approach involves rearranging the EEG channel vector to reflect their spatial configuration on the scalp before feeding them into a parallel RNN-based model comprising temporal LSTM and spatial BiLSTM. Alternatively, Luo et al. [56] proposed the use of FBCSP algorithm to extract spatial features of the EEG signal, which are then fed into a gated recurrent unit (GRU) for temporal feature learning. Similarly, Niu et al. [57] combined FBCSP with a convolutional LSTM (C-LSTM) network. Furthermore, hybrid models, such as attention-based DSC-ConvLSTM [58], CNN-GRU [59], and CNN-LSTM models with DWT decomposition [60], integrate CNN and LSTM to capture the spatial and temporal features of EEG signals, respectively.

However, a fundamental limitation of these models lies in their oversimplification of the spatial relationships among EEG channels. Specifically, they represent the spatial configuration of EEG electrodes as a linear structure, overlooking the non-Euclidean topology of EEG data and the functional interactions within the brain. In practice, EEG electrodes are systematically positioned across the scalp following standardized configurations (e.g., the 10-20 system), forming a network of interconnected nodes that reflects the underlying neural activity. The spatial proximity of the electrodes and their relative positions on the scalp play a critical role in capturing localized neural activity and propagation of neural signals [61]. Consequently, the inability of these architectures to precisely model the spatial dependencies of EEG channels significantly limits their effectiveness in distinguishing neural activities.

### 2) GRAPH-BASED SPATIOTEMPORAL MODELS

Meanwhile, graph-based spatiotemporal models leverage a graph structure to preserve the topographic configuration of EEG electrodes, facilitating a more accurate representation of the spatial relationships. Recent advancements have integrated graph neural networks (GNNs), which are specifically designed to process graph-structured data with nodes representing entities and edges capturing the relationships between them [62]. For instance, the graph-based CRAM (G-CRAM) model [63] employs a graph embedding to represent the spatial relationships among EEG channels based on their actual scalp positions and combines CNN with a recurrent attention network (RAN) to capture the temporal dependencies. Vivek et al. [31] introduced the ST-GNN model, which integrates CNN and graph convolutional network (GCN) with a trainable weighted adjacency matrix to learn the optimal relationships among EEG channels. On the other hand, the spiking multihead adaptive graph convolution and LSTM neural network (SGLNet) [32] combines spiking neural networks (SNNs), LSTM, and multihead adaptive GCN with an EEG channel adjacency graph constructed using k-nearest neighbor (KNN) algorithm. Furthermore, incorporating attention mechanisms, the dense multi-scale graph attention network (DMT-GAT) [64] integrates CNN with a graph attention network (GAT) and uses a fully-connected graph to represent the EEG channel relationships, whereas EEGAT [34] adaptively constructs the channel adjacency graph using the KNN algorithm.

While these graph-based spatiotemporal models enhance EEG signal representations by accurately capturing both spatial and temporal dependencies, some models generalize EEG channel relationships as a fully-connected graph or rely on trainable graph topologies, overlooking the inherent anatomical and functional interactions of the brain. In particular, EEG channels with similar letter designations (e.g., Cz and C1) are associated with the same cortical region, while EEG channels with common numerical designations (e.g., C3 and Cp3) correspond to the same lateral axis positioning.

Moreover, these models generally employ GNNs with CNNs or RNNs in a sequential manner, treating the spatial and temporal features of EEG signals as independent components. However, brain activity is inherently characterized by synchronized neural processes unfolding across functionally interconnected regions over time, forming dynamic patterns that span both spatial and temporal dimensions [65]. The decoupling of spatial and temporal feature learning in these models hinders their ability to capture intricate interdependencies between these dimensions [66]. This limitation is evident in numerous GNN-based models for time-series analysis, such as STGCN [67], DGCNN [68], and HGCN [69], which employ a "sandwich" architecture of spatial and temporal gating layers. Consequently, existing graph-based spatiotemporal MI-EEG models are constrained in effectively leveraging the rich spatiotemporal dependencies inherent in EEG signals.

## III. ATTENTIVE GRAPH-TEMPORAL CONVOLUTIONAL NETWORK

To address the limitations of existing works highlighted above, this study introduces the attentive graph-temporal convolutional network (AGTCNet), a novel spatiotemporal model that integrates GNNs and CNNs to effectively capture the spatiotemporal dynamics of multichannel EEG signals. In contrast to existing models that capture spatial and temporal features independently using separate and distinct deep learning architectures, AGTCNet jointly learns expressive spatiotemporal EEG representations while preserving the topology of EEG channels through a graph-based representation, enhancing the model's ability to generalize EEG signals across diverse users and time periods. Inspired by EEGNet and Transformer encoder architectures, AGTCNet consists of five core modules: channel-wise temporal convolution (CTC) module, graph convolutional attention network (GCAT) module, global convolutional adaptive pooling (GCAP) module, global temporal convolution (GTC) module, and temporal context enhancement (TCE) module.

As illustrated in Fig. 1, a key innovation of AGTCNet lies in the incorporation of an EEG channel adjacency graph, strategically constructed based on the anteroposterior (front-to-back) and mediolateral (side-to-side) topographic configuration of EEG electrodes, which serves as an inductive bias to enable effective spatial feature learning. Specifically, the GCAT leverages this adjacency graph to facilitate message passing among adjacent EEG channels, attentively capturing the spatiotemporal features of EEG signals and enhancing the robustness of the representations for each individual channel.

Building upon this foundation, the proposed AGTCNet processes two inputs: the multichannel EEG signal and an EEG channel adjacency graph that models the spatial relationships among EEG channels. The architecture begins with the CTC module, which transforms the raw EEG signal into more expressive representations suitable for downstream processing. This is followed by the GCAT module, which jointly captures the spatiotemporal dependencies of the EEG signal through an attentive message-passing mechanism among the individual channels. The GCAP module then adaptively aggregates the multichannel EEG representations into monochannel representations by learning the relevance weights of each EEG channel across multiple subspaces. Subsequently, the GTC module refines the monochannel EEG representations by capturing intricate temporal dynamics while reducing temporal dimensionality. Finally, the TCE module further enhances these EEG representations by capturing salient temporal features. Through the integration of these core modules, AGTCNet jointly learns both spatial (i.e., interchannel) and temporal dependencies, expressively transforming multichannel EEG signals into highly discriminative representations for classification tasks. The complete code is available in the AGTCNET repository published at https://github.com/galvinlim/AGTCNet.

### A. INPUT REPRESENTATION
Unlike most existing models that rely solely on multichannel EEG signals as input, AGTCNet incorporates an additional input—the EEG channel adjacency graph—facilitating the utilization of GNNs, specifically the GCAT.

#### 1) EEG SIGNAL
First, the model takes the multichannel EEG signal $X \in \mathbb{R}^{C \times T \times F}$, where $C$ represents the number of channels (i.e., nodes in GNNs), $T$ denotes the number of time points, and $F$ corresponds to the feature dimension. A sample 22-channel EEG signal is shown in Fig. 1. Throughout this paper, the notation ($C, T, F$) is consistently used, and the terms "EEG channel" and "node" are used interchangeably.

#### 2) EEG CHANNEL ADJACENCY GRAPH
Additionally, an EEG channel adjacency graph is defined and provided as input to the model, as depicted in Fig. 1. It is a static, undirected graph represented by a symmetric adjacency matrix $A \in \{0,1\}^{C \times C}$, where each node corresponds to an EEG channel, and the edges are bidirectional. Particularly, the graph is static, with fixed adjacent channels for each EEG channel across all time points. The adjacency matrix is defined as a binary matrix, where a value of 1 denotes connectivity between adjacent EEG channels based on the anteroposterior (i.e., same letter designation) and mediolateral (i.e., same numerical designation) topographic configuration of the electrodes on the scalp. Fig. 2 presents the EEG channel adjacency graph configurations for the BCICIV2A and EEGMMIDB datasets. This connectivity effectively captures the spatial correlations among EEG channels located within the same cortical region and along the same lateral axis.

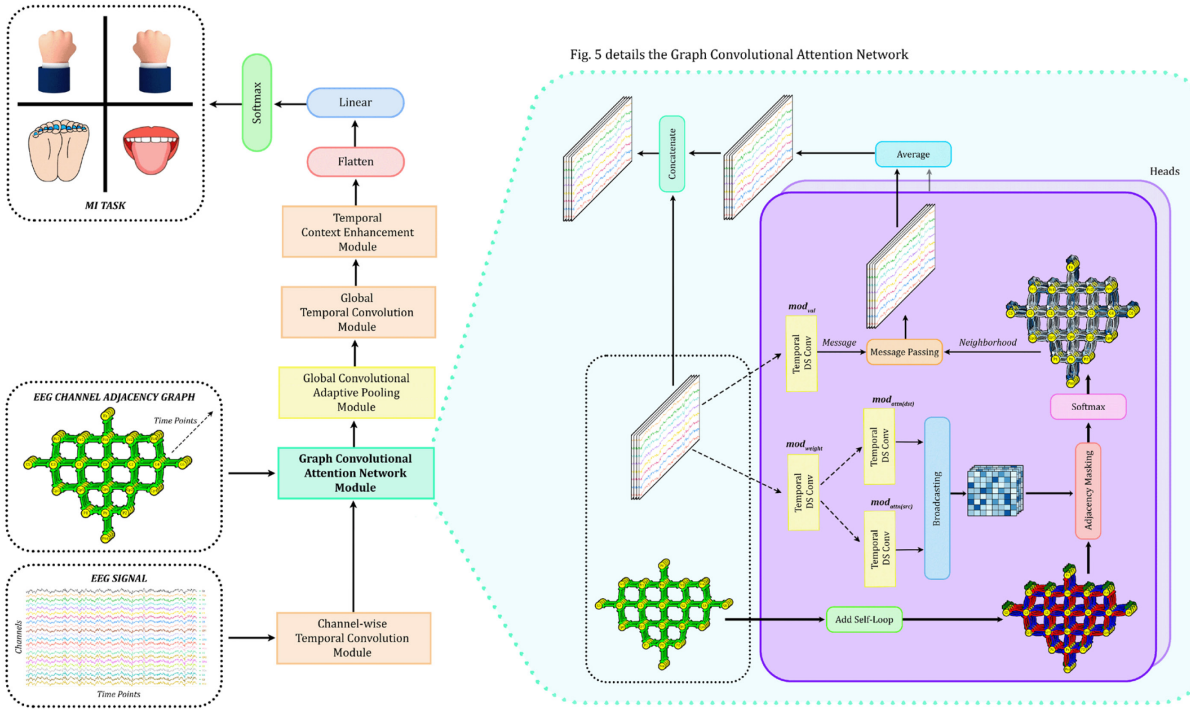

**FIGURE 1.** Attentive graph-temporal convolutional network architecture.

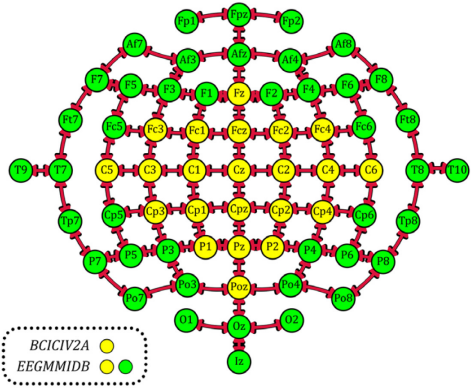

**FIGURE 2.** EEG channel adjacency graphs for BCICIV2A and EEGMMIDB datasets.

*B. PREPROCESSING*

In this study, preprocessing is applied to the EEG signals to eliminate irrelevant information, enhance signal robustness, and optimize the learning process of the model. However, given that EEG datasets were acquired using different EEG headsets with distinct signal specifications and underwent varying preliminary preprocessing procedures by their respective publishers, customized preprocessing pipelines are employed to ensure a standardized representation of the input EEG signals.

---
**Algorithm 1** EEG preprocessing
---
Assume EEG signal $X \in \mathbb{R}^{C \times T}$
**if** $X$ in $V$ unit **then**           ▷ apply scaling
  $X \leftarrow X \times 10^6$
**if** $f_s > 200\ Hz$ **then**           ▷ apply downsampling
  **for** $c \leftarrow 1, 2, \ldots, C$ **do**
    $x_c \leftarrow \text{ButterworthLPFilter}\left(x_c, \frac{1}{2} f_{s(new)}, \text{order} = 12\right)$
    $x_c \leftarrow \text{MNE\_Resample}\left(x_c, f_{s(new)}\right)$
$x_c(t) \leftarrow x_c(t) - \frac{1}{C}\sum_{i=1}^{C} x_i(t)$   ▷ apply common average referencing

---

*1) SCALING*

EEG signals are generally measured in the microvolt ($\mu V$) range, exhibiting small absolute magnitudes. To maintain consistency throughout the study, the EEG signals from the EEGMMIDB dataset are scaled by a factor of $1 \times 10^6$, following the preprocessing method applied to the BCICIV2A dataset.

*2) DOWNSAMPLING*

Afterward, downsampling is performed to enhance the model training efficiency by reducing data complexity while eliminating irrelevant high-frequency components. Specifically, for the BCICIV2A dataset, the sampling rate is halved from 250 Hz to 125 Hz to prevent interpolation artifacts and to preserve critical MI information within the 8 Hz to 30 Hz frequency band [5]. Notably, the sampling rate was not reduced to the Nyquist frequency of exactly 30 Hz to avoid the complete loss of potentially relevant information outside the theoretical MI frequency band [70], a decision supported by empirical tuning results that balanced the signal robustness and model complexity. To prevent aliasing, a 12th-order Butterworth low-pass filter is applied before resampling to eliminate high-frequency components above the new Nyquist frequency (i.e., 62.5 Hz). Resampling is then performed using the MNE-Python function, which employs the fast Fourier transform (FFT) method with a boxcar window function to preserve the integrity of the signal frequency characteristics and minimize spectral leakage. As for the EEGMMIDB dataset, no downsampling is performed, as the original sampling rate of 160 Hz (i.e., a Nyquist frequency of 80 Hz) was deemed appropriate based on the tuning results.

*3) COMMON AVERAGE REFERENCING*

Finally, common average referencing (CAR) is applied to both datasets, as employed in previous studies [31], [32], [71], [72]. This rereferencing method involves subtracting the average value of all EEG channels from each individual channel at every time point, as illustrated in Fig. 3. The rereferenced signal $x_c^{CAR}(t)$ for channel $c$ at time $t$ is defined as:

$$x_c^{CAR}(t) = x_c(t) - \frac{1}{C}\sum_{i=1}^{C} x_i(t) \qquad (1)$$

where $C$ denotes the total number of channels and $\frac{1}{C}\sum_{i=1}^{C} x_i(t)$ represents the average signal across all channels at time $t$. Specifically, the average is computed across 22 channels for the BCICIV2A dataset and 64 channels for the EEGMMIDB dataset, in accordance with their respective channel configurations. This process essentially rereferences the EEG signal to a common average, eliminating bias introduced by a single reference point on the scalp. As a result, the overall electrical activity (i.e., signal amplitude) across all channels sums to zero at each time point, localizing the channel information and enhancing the comparability of EEG signals across channels. Moreover, CAR effectively reduces uncorrelated signal sources and common noise (e.g., electrical interference) among the EEG channels, improving the signal-to-noise ratio (SNR) [73]. This preprocessing technique thus enables the model to better capture the underlying neural dynamics, facilitating effective feature learning.

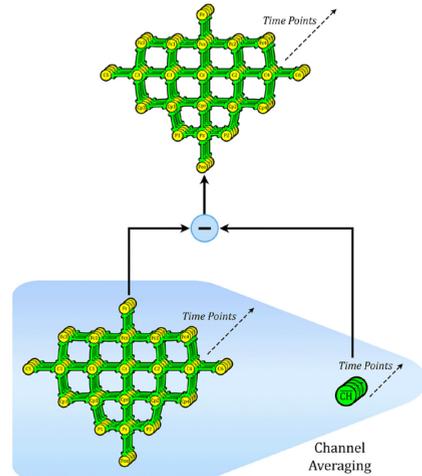

**FIGURE 3.** EEG common average referencing.

## C. CHANNEL-WISE TEMPORAL CONVOLUTION MODULE

Raw EEG signals with a feature size of 1 generally contain no significant information at individual time points. Hence, similar to EEGNet, the channel-wise temporal convolution (CTC) module is employed to transform EEG signals into more expressive representations, enabling effective message passing among adjacent EEG channels in the subsequent GCAT module.

Each EEG channel signal is initially passed through a common temporal convolutional layer with a kernel size of $(1, 32)$, stride of $(1, 1)$, and `valid` padding, as shown in Fig. 4. This convolutional layer essentially acts as a sliding window filter across the temporal dimension, capturing local temporal dependencies and transforming the multichannel EEG signal into high-level multichannel representations (i.e., $\mathcal{F} = 8$). Batch normalization (BN) is then applied to normalize the output, enhancing network stability.

Additionally, temporal average pooling with a pool size of $(1, 4)$ and stride of $(1, 2)$ is applied to each EEG channel, reducing the temporal dimension by approximately half. This operation not only minimizes the computational complexity of subsequent modules but also functions as an attention mechanism, emphasizing the most relevant features. In particular, a stride of half the pool size ensures a smooth transition between adjacent pooled segments, effectively preserving the long-term temporal dependencies.

## D. GRAPH CONVOLUTIONAL ATTENTION NETWORK MODULE

The transformed high-level multichannel EEG representations are then fed into the graph convolutional attention network (GCAT), a modified version of the graph attention network (GAT) [74], [75]. GCAT aggregates information from a node's neighbors through an attention mechanism that dynamically assigns weights based on the learned relevance of each neighboring node. This mechanism enables the model to focus on salient neighbors, facilitating the learning of expressive and context-aware node representations. However, unlike GAT, which uses linear layers, GCAT leverages convolutional layers to learn the node representations and the attention weights of each node with respect to its neighboring nodes and itself. This modification is motivated by the additional temporal dimension in the node features (i.e., EEG channel features); hence, temporal convolutional layers are introduced to capture the localized temporal dependencies of each EEG channel as message passing progresses. Instead of relying solely on its own information, GCAT enables EEG channels to aggregate local temporal features from adjacent channels, thereby enhancing the robustness of each EEG channel.

To further elaborate on the GCAT architecture, consider an EEG channel (i.e., node) $i$ with features $X_i$ of dimension $(\mathcal{C} = 1, \mathcal{T}, \mathcal{F})$ and adjacent channels (i.e., neighboring nodes) $j \in \mathcal{N}(i)$, each with features $X_j$ of the same dimension. The GCAT is mathematically expressed as:

$$X_i^* = \sum_{j \in \mathcal{N}(i)} \left( \alpha_{ij} \mathbf{1}_{\mathcal{F}^*}^\top \right) \otimes \sigma(W_{val} * X_j) \quad (2)$$

$$\alpha_{ij} = exp(e_{ij}) \oslash \sum_{k \in \mathcal{N}(i)} exp(e_{ik}) \quad (3)$$

$$e_{ij} = PReLU \left( \sigma(W_{src} * X_i) + \sigma(W_{dst} * X_j) \right) \quad (4)$$

where $\otimes$ denotes Hadamard product (i.e., elementwise product), $\oslash$ denotes Hadamard division (i.e., elementwise division), $*$ denotes the convolution operation, and $\sigma$ represents a nonlinear activation function.

In the actual implementation, as illustrated in Fig. 5, two temporal convolutional layers are employed to perform the source node temporal convolution $W_{src}$ and destination node temporal convolution $W_{dst}$ in the attention mechanism $\alpha \in [0, 1]^{\mathcal{C} \times \mathcal{C} \times \mathcal{T} \times 1}$, as defined below:

$$\sigma(W_{src} * X_i) := \sigma \left( W_{attn(src)} * \sigma(W_{weight} * X_i) \right) \quad (5)$$

$$\sigma(W_{dst} * X_j) := \sigma \left( W_{attn(dst)} * \sigma(W_{weight} * X_j) \right) \quad (6)$$

where a shared temporal convolutional layer $W_{weight}$ is first applied to the EEG channel features $X$, transforming them into higher-level representations, and distinct temporal convolutional layers $W_{attn(src)}$ and $W_{attn(dst)}$ are then employed to yield scalar outputs, representing the attention coefficients for the source and destination nodes, respectively, at each time point. These source and destination attention coefficients are broadcast and passed through a PReLU nonlinearity to form an unnormalized edge attention coefficient matrix $e \in \mathbb{R}^{\mathcal{C} \times \mathcal{C} \times \mathcal{T} \times 1}$, which represents a fully-connected graph. Unlike the Leaky ReLU activation function used in the original GAT [74], [75], which introduces a fixed negative slope $\alpha$ to address the "dying ReLU" problem in ReLU, Parametric ReLU (PReLU) employs a learnable parameter for $\alpha$, offering greater flexibility and adaptability in modeling complex relationships within dynamic node interactions, as validated in model tuning [76]. The resulting fully-connected graph allows each node to pass messages to all the other nodes. To impose the graph topology, adjacency masking with self-attention is applied, transforming the undirected graph into a weighted graph, as depicted in Fig. 6. Finally, the adjacency attention coefficients are normalized using the softmax function across all neighboring nodes $j \in \mathcal{N}(i)$ and passed through a dropout (DO) layer to enhance regularization and mitigate overfitting.

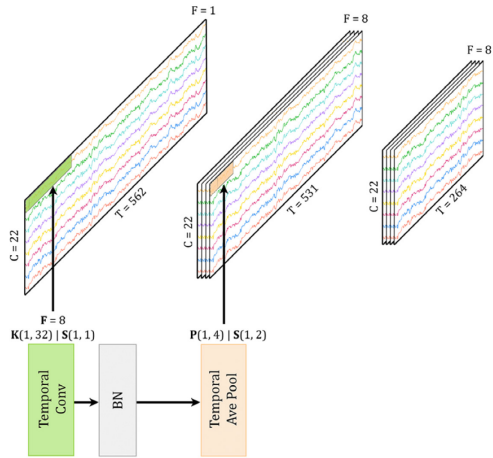

**FIGURE 4.** Channel-wise temporal convolution module.

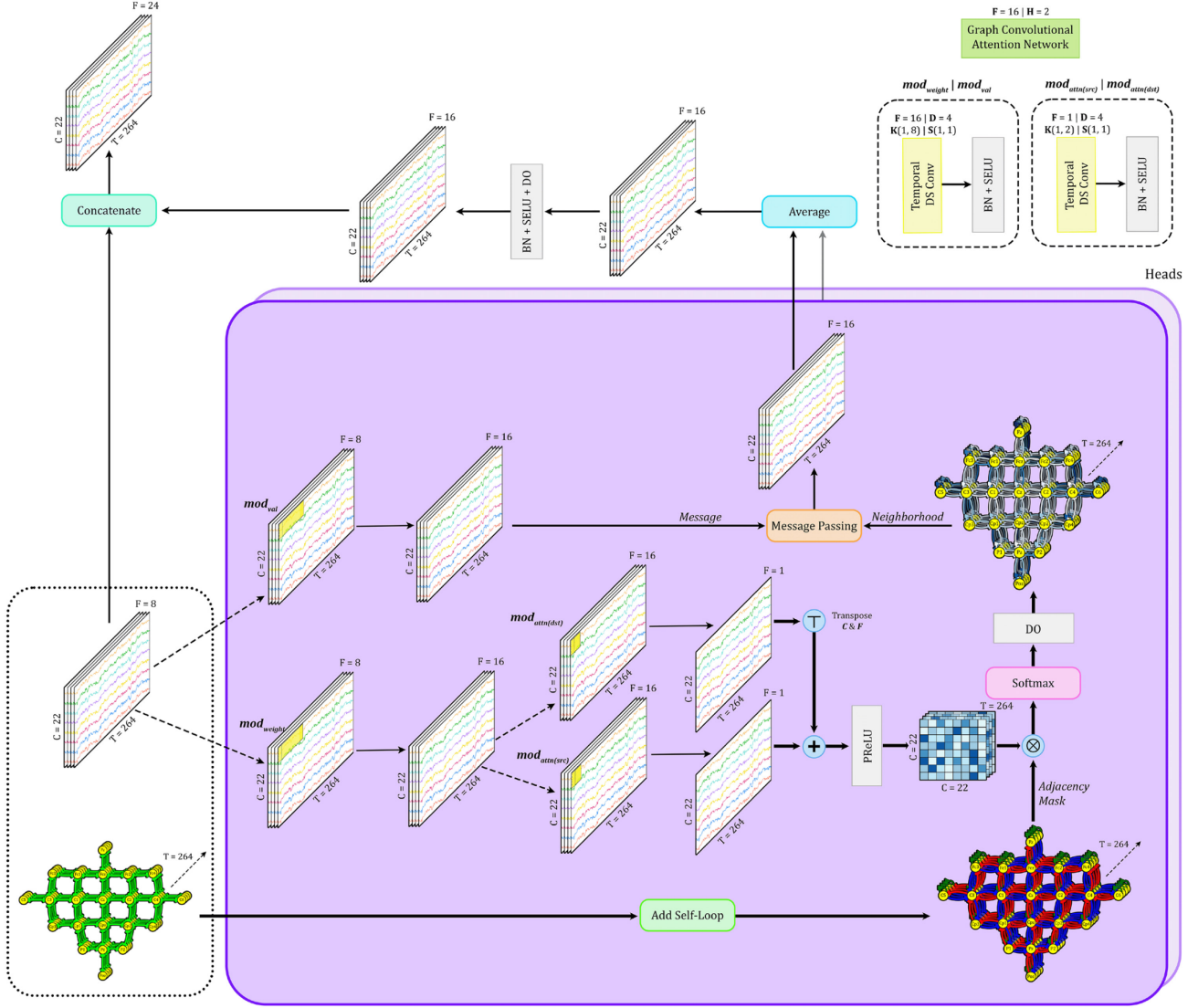

**FIGURE 5.** Graph convolutional attention network module.

For the actual node representations (i.e., messages) of each EEG channel $X_i$, a separate temporal convolutional layer $W_{val}$ is applied to capture node-specific temporal features. With the learned node representations and the weighted graph (i.e., normalized adjacency attention coefficient matrix), message passing is performed on each EEG channel by taking the linear combination of its features and those of its adjacent channels, weighted by the corresponding attention coefficients.

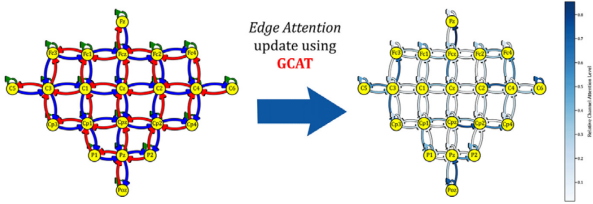

**FIGURE 6.** Update of attention weights on the EEG channel adjacency graph in GCAT.

Similar to GAT, multihead attention is employed with each head $h$ performing the transformation in (2). This mechanism enables the model to learn diverse representations from different subspaces, thereby stabilizing the learning process of self-attention. The outputs from all $H$ heads are then averaged to preserve the output feature size, as defined below:

$$X_i = \sigma \left( \frac{1}{H} \sum_{h=1}^{H} \sum_{j \in \mathcal{N}(i)} \left( \alpha_{ij}^{(h)} \mathbf{1}_{\mathcal{F}^*}^{\top} \right) \otimes \sigma \left( W_{val}^{(h)} * X_j \right) \right) \quad (7)$$

This is followed by batch normalization, a SELU nonlinearity, and a dropout layer. Finally, a residual connection is employed, where the input (0-hop features) and output (1-hop features) of GCAT are concatenated, allowing the model to retain both the original channel features and the aggregated features from adjacent EEG channels.

In the proposed AGTCNet, two attention heads are employed in GCAT with an output feature size of 16. Depthwise separable convolution (DS Conv) is primarily utilized due to its computational and parameter efficiency while maintaining comparable performance to that of standard convolutions. Specifically, the temporal convolutional layers $W_{weight}$ and $W_{val}$ utilize depthwise separable convolutions with a kernel size of $(1,8)$, stride of $(1,1)$, and depth of 4 to capture local temporal dependencies and transform the node features into higher-level representations (i.e., $\mathcal{F}=16$). The temporal convolutional layers $W_{attn(src)}$ and $W_{attn(dst)}$ then reduce the node features to a feature size of 1 using depthwise separable convolutions with a kernel size of $(1,2)$, stride of $(1,1)$, and depth of 4. Following each convolutional layer, the outputs are batch normalized and passed through the SELU nonlinearity.

Overall, the GCAT module facilitates informative message passing among adjacent EEG channels, enabling the joint capture of spatiotemporal features from the EEG signal. Particularly, a 1-hop GCAT is employed to aggregate features between immediate adjacent channels, capturing the salient local spatial features among the EEG channels. Moreover, to mitigate oversmoothing, where feature aggregations from broader subgraphs lead to the loss of local information and result in homogeneous node features [77], a residual connection is employed to preserve the original channel features, ensuring that critical local information is retained. This design is consistent with the tuning results, which reveal that 1-hop GCAT with a residual connection offers an effective balance between capturing spatial dependencies among EEG channels and preventing oversmoothing.

Furthermore, instead of employing fully-connected or trainable graph topologies, the EEG channel adjacency graph is strategically constructed based on the actual topographic configuration of EEG electrodes on the scalp, introducing an inductive bias into the model while reducing its complexity. By leveraging the spatial correlations and functional relationships among EEG channels [61], the predefined graph topology effectively eliminates spurious connections between spatially distant and functionally uncorrelated channels, thereby enhancing interpretability and simplifying the model's learning process.

### E. GLOBAL CONVOLUTIONAL ADAPTIVE POOLING MODULE

Adapted from EEGNet, the multichannel EEG representations are then aggregated into monochannel EEG representations using the global convolutional adaptive pooling (GCAP), as illustrated in Fig. 7. Given the functional localization of MI-related neural activity within the brain, GCAP is employed to adaptively focus on specific EEG channels by learning their signed relevance weights, unlike traditional pooling methods, which utilize fixed statistical techniques, such as mean and sum. GCAP essentially functions as learnable spatial filters, as depicted in Fig. 8, that can be trained in an end-to-end manner, allowing varying filter weights for each subspace. This contrasts with CSP, which employs fixed spatial filters that produce filtered signals irrespective of the downstream task. In this module, spatial depthwise convolution (DW Conv) with a kernel size corresponding to the number of EEG channels (i.e., $(22,1)$ for BCICIV2A and $(64,1)$ for EEGMMIDB) and `valid` padding is applied at each time point, followed by batch normalization and a SELU nonlinearity.

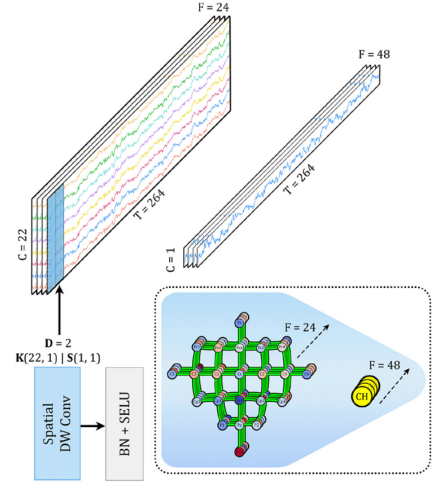

**FIGURE 7.** Global convolutional adaptive pooling module.

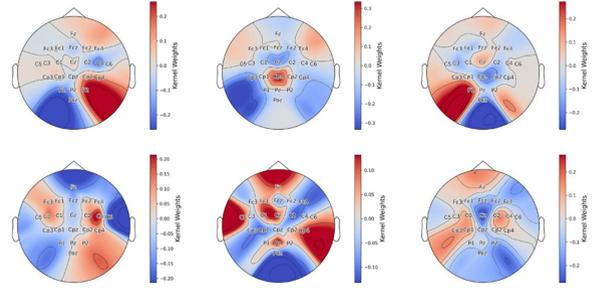

**FIGURE 8.** Sample kernel filters in GCAP.

### F. GLOBAL TEMPORAL CONVOLUTION MODULE

With the learned monochannel EEG representations, a series of temporal pooling and convolution operations is applied through the global temporal convolution (GTC) module, further transforming them into higher-level representations by capturing intricate temporal dynamics while reducing the temporal dimension. As shown in Fig. 9, the EEG representations first undergo temporal average pooling with a pool size of $(1,4)$ and stride of $(1,4)$, reducing the number of time points by a factor of 4, followed by a dropout layer. The pooled representations are then processed through a temporal depthwise separable convolution with a kernel size of $(1,8)$, stride of $(1,1)$, and depth of 4, capturing higher-level temporal features (i.e., $\mathcal{F}=96$), followed by batch normalization and a SELU nonlinearity. A second temporal average pooling, identical to the first, further reduces the temporal dimension by another factor of 4. Thus, the GTC module effectively doubles the feature dimension while compressing the temporal dimension by approximately a factor of 16, resulting in compressed high-level EEG representations.

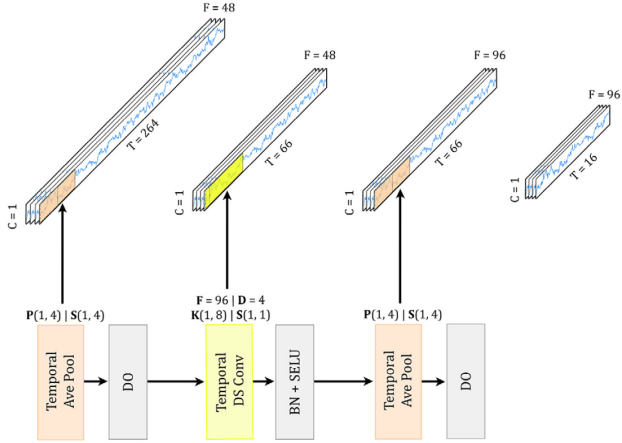

**FIGURE 9.** Global temporal convolution module.

### G. TEMPORAL CONTEXT ENHANCEMENT MODULE

The temporal context enhancement (TCE) module, inspired by the Transformer encoder block [78], further refines the monochannel EEG representations by learning the critical temporal features embedded within them. As illustrated in Fig. 10, the EEG representations are first augmented with positional encoding to preserve their temporal dependencies prior to being flattened for final classification. Specifically, the positional encoder embeds positional information using sine and cosine waves, enabling the neural network to capture the temporal aspects of the EEG representations, even in the absence of the temporal dimension.

Moreover, given the time-varying nature of MI task execution, where different subjects and trials exhibit varying delays, a multihead attention (MHA) layer is employed to dynamically focus on the salient segments of the EEG sequence and capture long-range temporal dependencies. This attention mechanism effectively equips the model with time invariance, facilitating the identification of MI stimuli across the entire input EEG signal. Specifically, two attention heads are used in the MHA layer, with the dimensions of key, query, and value set to 8. Batch normalization and dropout are then applied, and the resulting output is summed with the original input fed into the MHA layer.

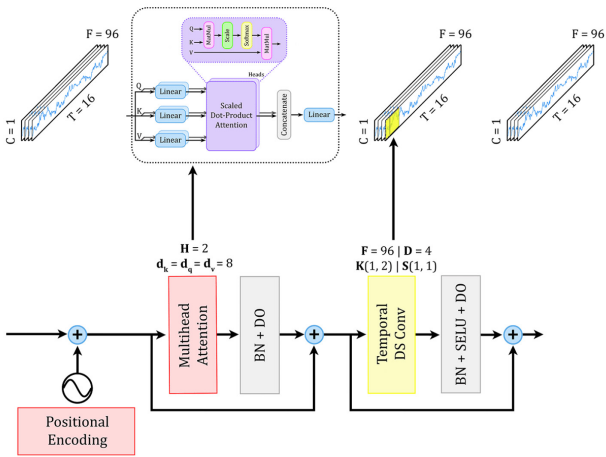

**FIGURE 10.** Temporal context enhancement module.

Following this, a final temporal feature learning is performed using temporal depthwise separable convolution with a kernel size of $(1, 2)$, stride of $(1, 1)$, depth of 4, and output feature size of 96. A convolutional layer is specifically utilized for the feedforward network, instead of a linear layer, to introduce locality within the module [79], as demonstrated by its enhanced performance in model tuning while maintaining parameter efficiency. Batch normalization, SELU nonlinearity, and dropout are then applied, and the resulting output is again summed with the original input fed into the temporal convolution layer. In particular, residual connections are employed in both the MHA and temporal depthwise separable convolution layers to preserve the original input features while enhancing the flow of information by maintaining a direct path for gradient propagation.

### H. FINAL CLASSIFICATION

Finally, the compressed high-level EEG representations are flattened and passed through a linear layer with an output size corresponding to the number of MI tasks being classified, as shown in Fig. 1. A softmax activation function is then applied for the final classification.

### I. MODEL TUNING FRAMEWORK

The proposed AGTCNet is developed through extensive experimentation. Specifically, model tuning was conducted using Optuna, which employs Bayesian optimization through the tree-structured Parzen estimator (TPE) algorithm to efficiently explore the optimal set of hyperparameters within the defined search space [80], [81]. By leveraging Bayesian statistics, Optuna assesses the performance of previous hyperparameter combinations to inform the selection of the potential best set of configurations for subsequent iterations. More information is provided in APPENDIX A.

## IV. PRACTICAL EEG MODEL TRAINING AND EVALUATION FRAMEWORK

In model evaluation, it is essential that the assessment is representative of real-world deployment conditions. Ideally, an EEG classifier should be subject-independent, capable of capturing subject-invariant EEG representations while maintaining a robust performance across diverse users. However, due to the current limitations of EEG technology and the inherent complexity of the human brain, achieving a satisfactory performance remains a significant challenge. Consequently, numerous studies have focused on subject-specific (subject-dependent) EEG classifiers that are capable of capturing session-invariant EEG representations tailored to individual users, often yielding superior performance.

Generally, model performance is evaluated on a new dataset that is not seen during the training phase. Data leakage is a modeling flaw wherein information from the validation or test dataset inadvertently influences the training process, resulting in overly optimistic performance estimates and poor generalization. Consequently, the model may appear to perform well during offline testing but fails upon real-time deployment. This issue typically arises when data preprocessing and feature extraction are performed on the entire dataset or when model performance is reported based on the validation dataset used to guide optimal model decisions, such as model checkpointing and early stopping [82]. To address this, it is standard practice to perform train-validation-test splits in model evaluation, splitting the dataset into three mutually exclusive subdatasets: training, validation, and test datasets. The training dataset is mainly used to optimize the model parameters, while the validation dataset informs model training decisions. Finally, the selected best model is evaluated on the test dataset to provide generalized performance estimates [82].

However, this evaluation framework is only feasible for large datasets. EEG datasets, which are constrained by the time-consuming and strenuous data collection process from subjects, are often limited in size. This limitation renders the train-validation-test split evaluation framework impractical, as it further reduces the already limited samples available for model training and evaluation, compromising the model's generalizability and robust estimation of model performance. Hence, most EEG studies only perform a train-validation split, which is also adopted in this study to enable the benchmarking of the proposed model with other state-of-the-art models. Since it only involves two subdatasets, the validation dataset is often interchangeable with the test dataset. To address the limitations of the train-validation split evaluation framework and mitigate the risk of performance overestimation, a new performance metric is introduced and utilized. Furthermore, cross-validation (CV) is performed by repeatedly splitting the dataset and using different random initializations to account for stochasticity in the model training. The arithmetic mean of the model's performance across all validation sets is then reported, offering a robust estimate of its generalizability [82].

Given the complex and dynamic nature of the brain, it is imperative to ensure that model training and validation comprehensively account for intersubject and intersession variability. Hence, this study formally establishes model training-evaluation frameworks for EEG classifier development.

### A. SUBJECT-INVARIANT ASSESSMENT

Considering the brain's intricate and variable nature across individuals, the development of a subject-invariant (subject-independent) EEG classifier is essential. To assess the model's ability to learn subject-invariant representations of EEG signals and accurately classify them across different subjects, three possible training-evaluation frameworks are identified: (1) subject-level (SL), (2) subject-mixed (SM), and (3) subject-mixed with new validation subject/s (SN). In the SL framework, the model is trained and validated on a single subject. In contrast, the SM and SN frameworks involve training on multiple subjects, with a distinction in the validation dataset. Specifically, the SM framework evaluates the model on the same set of subjects used for training, whereas the SN framework tests the model on a new subject or a disjoint set of subjects excluded from the training dataset. The SN framework can be performed using leave-one-subject-out (LOSO) CV, a widely used method in studies utilizing the BCICIV2A dataset [29], [30], or leave-multiple-subjects-out (LMSO) CV for large datasets, such as the EEGMMIDB dataset [32]. In LOSO-CV, one subject is set aside for validation, while the remaining subjects are used for model training, resulting in $S$ distinct training and validation dataset splits for $S$ subjects. On the other hand, in LMSO-CV, the $S$ subjects are divided into $k$ mutually exclusive subsets through $k$-fold CV, where $k-1$ subsets are utilized for model training and the remaining subset serves as the validation dataset, yielding $k$ distinct training and validation dataset splits.

The rationale behind these frameworks is that the SL framework leads to a model biased toward the neural dynamics of a particular subject, with its validation performance ensuring functionality solely for that subject, thus resulting in a subject-specific classifier. The SM and SN frameworks, on the other hand, enable the model to learn neural dynamics across different subjects, capturing subject-invariant EEG representations. However, as employed in some previous studies [30], [32], the SM framework only ensures functionality for a particular set of subjects and fails to generalize across all users. Hence, the SN framework is considered the most practical model training-evaluation framework, as its validation ensures functionality for general users, specifically subjects not involved in the model training, thus aligning with the goal of making BCI technology accessible to the public without the need for user calibration.

### B. SESSION-INVARIANT ASSESSMENT

Another crucial consideration of the human brain is its continuously evolving temporal neural dynamics, which change over time depending on one's present mental state, emotional condition, and external influences, underscoring the necessity of developing a session-invariant (subject-specific) EEG classifier. Given that EEG datasets typically consist of multiple recording sessions conducted at different time periods (e.g., on different days), these sessions inherently exhibit variability in temporal neural dynamics. To assess the model's ability to learn session-invariant representations of EEG signals and accurately classify them across different time periods, two possible training-evaluation frameworks are identified: (1) distinct train-validation session (DS) and (2) random train-validation session (RS). In the DS framework, distinct sessions are explicitly selected as training and validation datasets using leave-one-session-out (LOSeO) CV. In contrast, the RS framework performs random CV, where all sessions are combined and randomly stratified into training and validation datasets, ensuring balanced splits.

The rationale behind these frameworks is that the RS framework, as employed in some studies [83], [84], fails to assess the model performance on EEG signals recorded at different time periods. Moreover, improper implementation of this framework introduces the risk of data leakage, rendering the model undeployable. For instance, as illustrated in Fig. 11, employing a sliding window cropping (SWC) strategy with overlapping segments followed by random train-validation split results in indistinct training and validation datasets, undermining the integrity of the validation process. This issue has often been overlooked in studies employing the SWC strategy for data augmentation [85], [86], [87]. Hence, the DS framework is deemed more practical, as it validates EEG signals recorded across different time periods, thus ensuring the model's functionality during deployment.

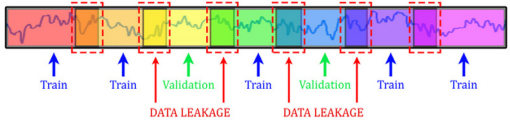

**FIGURE 11.** Illustration of data leakage introduced when sliding window cropping with overlapping segments is followed by random train-validation split.

### C. SYNTHESIS

Accounting for these two assessments, five distinct training-evaluation frameworks are formulated, as presented in Fig. 12: (1) subject-level with distinct train-validation session (SL-DS), (2) subject-level with random train-validation session (SL-RS), (3) subject-mixed with distinct train-validation session (SM-DS), (4) subject-mixed with random train-validation session (SM-RS), and (5) subject-mixed with new validation subject/s (SN). Notably, the SN framework comprises a single configuration, as the training and validation datasets involve different sets of subjects, eliminating the need for session selection. Additionally, considering the limited EEG dataset size, a fine-tuning (FT) strategy is introduced for the SL-DS framework. Specifically, the model is initially trained under the SN framework with the subject of interest as part of the validation dataset and subsequently fine-tuned under the SL-DS framework, resulting in the sixth training-evaluation framework: (6) subject-level with distinct train-validation session fine-tuned (SL-DS-FT).

Among these six training-evaluation frameworks, the SL-DS and SL-DS-FT frameworks, which account for intersession variability within a particular subject, are considered practical for developing a subject-specific EEG classifier. Meanwhile, the SN framework, which addresses intersubject variability while implicitly accounting for intersession variability across multiple subjects, is deemed most appropriate for developing a subject-independent EEG classifier. Given the limited samples per subject, this study employed the SL-DS-FT training-evaluation framework for subject-specific classification to maximize the utilization of available EEG data and further enhance model performance.

### D. PERFORMANCE METRICS

In previous EEG studies, accuracy and Cohen's kappa (aka kappa coefficient) are the primary performance metrics for evaluating EEG classification models. Accuracy, a standard metric for classification tasks, is defined as:

$$Acc = \frac{1}{N}\sum_{c=1}^{C} TP_c \quad (8)$$

where $TP_c$ denotes the true positive count (i.e., the number of correctly predicted samples) for class $c$ and $N$ is the total number of samples. Meanwhile, Cohen's kappa ($\kappa$) is a statistical measure that evaluates the level of agreement by the classifier, accounting for the likelihood of agreement occurring by chance, and is defined as:

$$\kappa_{score} = \frac{1}{N}\sum_{a=1}^{N} \frac{P_a - P_e}{1 - P_e} \quad (9)$$

where $P_a$ denotes the actual percentage of agreement, $P_e$ denotes the expected percentage of agreement by chance, and $N$ is the total number of samples. The $\kappa$-score ranges from -1 to 1, with a value of 0 indicating no agreement (i.e., random chance) and 1 indicating perfect agreement. A negative $\kappa$-score signifies disagreement or performance worse than expected by chance. Notably, Cohen's kappa is particularly suitable for imbalanced datasets [88]. However, given the typical EEG recording paradigms, EEG datasets are generally balanced, eliminating the risk of bias toward any particular class due to class frequency disparities.

In model training, it is common practice to employ early stopping and select the best-performing model based on validation accuracy over the entire training epoch. However, this strategy has a caveat, as it is only appropriate when using the train-validation-test split evaluation framework. In EEG studies, which utilize a train-validation split evaluation framework, accuracy and Cohen's kappa metrics can lead to an overly optimistic performance estimate due to data leakage, as highlighted in [82]. This issue is particularly evident when the model's validation curve exhibits significant fluctuations. Relying solely on the peak validation accuracy provides a misleading estimate of the model's generalizability, as it masks potential performance degradation when exposed to unseen data.

To address the limitations associated with the train-validation split evaluation framework and account for fluctuations in the model's learning curve, the moving average (MA) strategy, a method also utilized in TensorBoard, is introduced. This strategy provides a more robust measure by smoothing out short-term fluctuations (i.e., transient spikes) in the learning curve, thereby capturing the underlying model's performance trends. Specifically, this study proposes the use of a simple moving average (SMA) to evaluate the model's accuracy and loss, as defined below:

$$SMA_e = \frac{1}{n} \sum_{i=e-n+1}^{e} M_i \quad (10)$$

where $e$ represents the current epoch, $n$ is the window size (i.e., the number of consecutive epochs over which the running average is computed), and $M_i$ denotes the performance metric (e.g., accuracy or loss) at epoch $i$. This smoothing technique assigns equal weights to all data points within the window, making it suitable for identifying long-term trends without being significantly influenced by short-term fluctuations, in contrast to the exponential moving average (EMA). For the evaluation conducted in this study, a window size of 20 epochs was selected to balance between responsiveness and stability.

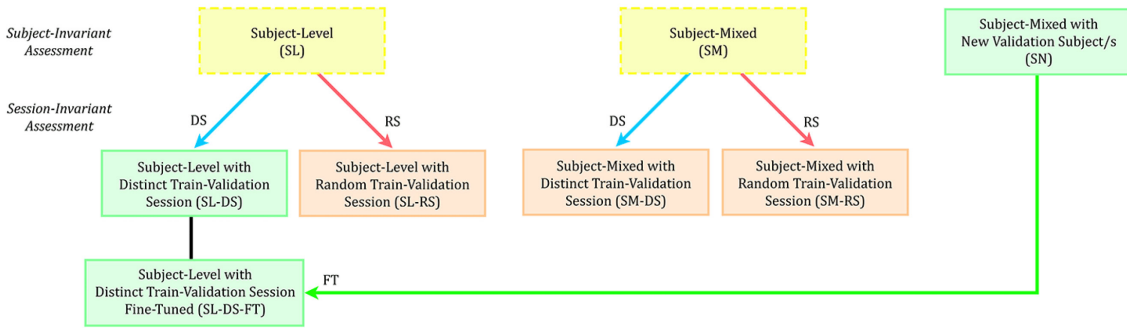

**FIGURE 12.** Practical EEG model training and evaluation framework.

In contrast to simply reporting the peak validation accuracy, evaluating model performance based on the maximum moving average validation accuracy across the training epochs provides a more reliable indicator of the model's ability to generalize to unseen data, as it mitigates the influence of transient performance peaks associated with the stochastic nature of model training. Furthermore, since convergence in model training is characterized by a plateau (i.e., minimal fluctuations) in the learning curve [89], [90], the moving average accuracy is expected to closely approximate the standard accuracy once the model has converged. Hence, this study deems the moving average accuracy metric to be a more practical and robust measure of model performance, as it increases confidence in the reported accuracy. Nevertheless, despite their susceptibility to performance overestimation, both standard accuracy and Cohen's kappa were also utilized in this study to enable benchmarking with previous studies, as these are commonly reported metrics.

## V. METHODOLOGY
### A. DATASET
This study extensively evaluated the proposed model using two Internet MI-EEG datasets: BCI Competition IV Dataset 2a (BCICIV2A) and EEG Motor Movement/Imagery Dataset (EEGMMIDB). Particularly, the BCICIV2A dataset was utilized for model development and tuning, and the model performance was comprehensively evaluated on both datasets. The selection of these Internet datasets was guided by several criteria, including the number of citations (spanning from 2004 to 2020, as reviewed in [91]) and dataset attributes such as signal quality, EEG channels, sampling rate, subject count, sample size per class for each subject, and MI tasks involved.

#### 1) BCI COMPETITION IV DATASET 2A
The BCICIV2A dataset [92] is a widely recognized benchmark dataset for MI-EEG studies [91]. EEG signals were recorded using 22 Ag/AgCl electrodes and three electrooculography (EOG) channels (excluded in this study). The signals were sampled at 250 Hz and subjected to bandpass filtering between 0.5 Hz and 100 Hz, with a 50-Hz notch filter applied to suppress power line noise. Amplifier sensitivity was set to 1 mV. The dataset contains four MI tasks: left hand, right hand, feet, and tongue, collected from nine subjects. Each subject participated in two recording sessions on separate days, with three baseline recordings and six MI runs per session. Specifically, during each MI run, 12 trials were performed for every MI task using a cue-based paradigm, with each trial lasting 4.0 seconds, resulting in 288 trials per session. In MI-EEG analysis, the signal timeframe typically spans 0.5 seconds prior to cue onset, capturing preparatory neural activity, and extends through the entire MI phase (i.e., -0.5 s to 4.0 s) for a total duration of 4.5 seconds [29], [30].

#### 2) PHYSIONET EEG MOTOR MOVEMENT/IMAGERY DATASET
The EEGMMIDB dataset [93], available on PhysioNet [94], is also extensively utilized in studies due to its substantial pool of subjects, making it suitable for developing a subject-independent MI-EEG classifier. EEG signals were acquired using a 64-channel EEG configuration at a sampling rate of 160 Hz on the BCI2000 system. The dataset comprises four motor tasks: left fist, right fist, both fists, and both feet, each involving either motor execution (ME) or motor imagery, recorded from 109 subjects. Each subject participated in a recording session of 14 runs: two baseline, six ME, and six MI runs. Specifically, using a cue-based paradigm, each motor run involved two motor tasks with around 15 trials, each lasting 4.0 seconds, followed by a 4.0-second break between trials. The typical signal timeframe used in analysis is restricted to the active MI phase (i.e., 0.0 s to 4.0 s) with a total duration of 4.0 seconds [30], [32]. Particularly, subjects 38, 88, 89, 92, 100, and 104 were excluded due to annotation errors [41], [95]; hence, only 103 subjects were utilized in this study.

### B. TRAINING PROCEDURE
This study was conducted using Python 3.11.0, TensorFlow 2.15.0, and CUDA 12.3 on a system running Ubuntu 22.04 LTS with an Nvidia Quadro RTX 6000 GPU (24 GB memory), an AMD EPYC 7402P CPU, and 256 GB of RAM. Additionally, Spektral 1.3.1 was utilized to implement GNNs.

The proposed model was trained for both 4-class and 2-class (Left Fist and Right Fist) MI-EEG classification tasks using categorical cross-entropy (CCE) and binary cross-entropy (BCE) loss functions, respectively. Adam optimizer was employed with an initial learning rate of $1 \times 10^{-3}$ ($5 \times 10^{-4}$ for fine-tuning), along with a learning rate (LR) decay scheduler. Specifically, the ReduceLROnPlateau scheduler was applied based on the validation loss with an LR decay factor of 0.9, patience of 10, cooldown of 0, and minimum learning rate set to $1 \times 10^{-4}$. This strategy facilitates model convergence by initially employing a high learning rate to accelerate the training process and escape suboptimal minima. The learning rate then gradually decays as training progresses, enabling the model to fine-tune its parameters and converge smoothly toward an optimal solution, thereby stabilizing the training process and mitigating overfitting [96].

Model parameters were initialized using the Glorot uniform initializer. For the BCICIV2A dataset, the model was trained for 1000 epochs with a batch size of 32 and early stopping patience of 300 epochs, while for the EEGMMIDB dataset, training was conducted for 500 epochs with a batch size of 32 and early stopping patience of 100 epochs. The early stopping strategy was employed to terminate model training when no further improvement in the validation accuracy was observed, thereby preventing overfitting and conserving computational resources.

### C. EVALUATION METHOD

The proposed model was evaluated for both subject-independent and subject-specific classifications, specifically under the SN and SL-DS-FT training-evaluation frameworks established in Section IV, respectively. To enhance the generalization of the model performance while accounting for its variability, cross-validation was performed by repeatedly splitting the dataset and training the model with different random initializations. For the SN training-evaluation framework, LOSO CV was applied to the BCICIV2A dataset, whereas LMSO CV, specifically nonrandom 5-fold CV (equivalent to an 80:20 train-validation split), was utilized for the EEGMMIDB dataset. The top-performing models under the SN training-evaluation framework were then utilized as the baseline models for the SL-DS-FT training-evaluation framework, employing LOSeO CV for both datasets. The number of training trials for each fold is specified in Table 1, with fewer samples performed for the EEGMMIDB dataset due to its large size, which is computationally expensive. The performance results were then averaged arithmetically across all folds and training samples to provide a generalized estimate of model performance.

Table 1 also summarizes the dataset specifications, input configurations, training hyperparameters, and cross-validation strategies for the BCICIV2A and EEGMMIDB datasets. Since these datasets were acquired from different EEG headsets with varying channel configurations, appropriate modifications were made to the AGTCNet parameters, particularly in the GCAP module, to ensure compatibility. While the proposed model was primarily developed using the BCICIV2A dataset with an EEG signal timeframe spanning from -0.5 s to 4.0 s—consistent with the typical signal timeframe adopted in previous studies—this study also investigated a signal timeframe confined to the active MI phase (i.e., 0.0 s to 3.0 s), as used in the EEGMMIDB dataset.

Finally, the proposed model was compared with state-of-the-art models from recent studies published between 2018 and 2024, with a particular focus on end-to-end deep learning and graph-based approaches utilizing raw EEG signals with minimal preprocessing to ensure a fair comparison. Studies that involved manual artifact removal were excluded due to its impracticality for real-time deployment, as well as those utilizing the SWC strategy and other data augmentation techniques due to concerns regarding potential data leakage.

**TABLE 1.** Specifications and training-evaluation methodologies for the BCICIV2A and EEGMMIDB datasets.

| | DATASETS | BCICIV2A | EEGMMIDB |
|---|---|---|---|
| **Dataset Specifications** | Subjects | 9 | 103 |
| | Sessions | 2 | 3 [a] |
| | Trials per Session | 288 | 28~32 |
| | Classes (MI Tasks) | 4 (Left Hand, Right Hand, Feet, Tongue) | 4 (Left Fist, Right Fist, Both Fists, Both Feet) |
| | Trial Paradigm | 2-s Fixation Cross 4-s MI Stimuli 2-s Rest | ~4-s Rest ~4-s MI Stimuli |
| **Input Configurations** | Signal Timeframe | Baseline: -0.5 s – 4.0 s (4.5 s) Proposed: 0.0 s – 3.0 s (3.0 s) | 0.0 s – 4.0 s (4.0 s) |
| | Sampling Rate | 250 Hz | 160 Hz |
| | EEG Channels | 22 | 64 |
| | Preprocessing | Downsampling + CAR | Scaling + CAR |
| **Training Hyper-parameters** | Batch Size | 32 | 32 |
| | Max Epochs | 1000 | 500 |
| | Early Stopping Patience | 300 | 100 |
| **Cross-Validations [b]** | Subject-Independent Classification | SN: LOSO CV (10 samples) | SN: LMSO CV (3 samples) |
| | Subject-Specific Classification | SL-DS-FT: LOSeO CV (5 samples) | SL-DS-FT: LOSeO CV (1 sample) |

[a] For the EEGMMIDB dataset, runs are considered as sessions.
[b] Samples refer to the number of training trials conducted for each CV fold.

Only studies with detailed evaluation methods, adhering to the practical model training-evaluation frameworks, were considered. Benchmarking was primarily conducted against studies utilizing the same dataset, number of classes, and evaluation metrics.

## VI. RESULTS AND DISCUSSION

### A. ABLATION STUDY

This section presents an ablation study of the proposed AGTCNet architecture, as well as the EEG preprocessing and model training methodologies. In particular, it comprehensively examines the impact of removing specific modules and processes using the BCICIV2A dataset. Table 2 summarizes the performance of AGTCNet, specifically the moving average accuracy (MA Acc) and accuracy (Acc), under various modifications. Generally, a notable performance improvement is observed when fine-tuning was employed for subject-specific classification. This fine-tuning strategy leverages the model's initial learning of the fundamental neural dynamics associated with MI stimuli across diverse subjects and further refines these learned EEG representations to adapt to the nuanced dynamics of the target subject.

Notably, the results show that training without LR decay significantly increased accuracy by up to 1.49%. However, it is evident from the model learning curves in Fig. 13 that this results in substantial fluctuations, which are undesirable compared to the smoother and more stable learning curve achieved with LR decay. As previously discussed, the goal of model training is to achieve convergence within the parameter space and to reach a plateau in the learning curve. The peaks observed in the fluctuating learning curve suggest stochastic improvements, rather than robust learning. Such fluctuations typically arise from a high learning rate, which allows the model to explore a wide parameter space and prevents it from converging to a local optimum. While this strategy increases the likelihood of finding high-accuracy parameter subspaces, it often leads to instability in the learning process, which may result in a model that fails to generalize to unseen data. In contrast, although LR decay limits the exploration of the parameter space and generally results in marginally lower peak accuracy, it is a standard practice in model training, as it facilitates stable learning, mitigates overfitting, and yields a robust model [97]. Compared to a model that achieves a higher peak accuracy but subsequently experiences a significant drop, the consistent performance improvement facilitated by LR decay suggests better generalization to unseen data. Thus, the moving average performance metric introduced in this study offers a more practical evaluation, as it accounts for fluctuations in the learning curve. Particularly, by smoothing out these fluctuations, it emphasizes the underlying trend of the model's learning performance, providing a more conservative and reliable performance estimate, as opposed to evaluating performance at a specific epoch.

TABLE 2. Performance of AGTCNet for subject-independent and subject-specific classifications on the BCICIV2A dataset under various modifications.

| Modifications | Subject-Independent | | Subject-Specific | |
|---|---|---|---|---|
| | MA Acc (%) | Acc (%) | MA Acc (%) | Acc (%) |
| Baseline: AGTCNet (-0.5 s – 4.0 s Signal) | 66.64 ± 9.99 | 70.06 ± 9.80 | 81.59 ± 7.65 | 82.75 ± 7.43 |
| No LR Decay | 65.67 ± 10.54 | **70.86 ± 9.98** | 82.18 ± 8.12 | **84.24 ± 7.24** |
| No CAR | 64.04 ± 8.91 | 67.78 ± 8.94 | 78.97 ± 9.19 | 80.57 ± 8.68 |
| No Downsampling (250 Hz) | 64.65 ± 10.95 | 68.63 ± 10.30 | 81.37 ± 7.32 | 82.55 ± 7.09 |
| No GCAT | 64.42 ± 10.04 | 67.39 ± 9.42 | 77.89 ± 9.53 | 79.04 ± 9.44 |
| No TCE | 65.55 ± 11.30 | 68.93 ± 10.98 | 79.24 ± 9.30 | 80.78 ± 8.64 |
| No GCAT & TCE | 63.80 ± 11.35 | 66.71 ± 10.53 | 77.42 ± 9.95 | 78.60 ± 9.80 |
| GCAT → GAT | 65.02 ± 10.30 | 68.72 ± 9.94 | 80.13 ± 8.33 | 81.37 ± 8.17 |
| GCAT: No Residual | 64.50 ± 11.39 | 67.60 ± 11.22 | 78.28 ± 9.83 | 79.78 ± 9.23 |
| CTC: No Pooling | 64.20 ± 11.62 | 68.45 ± 10.69 | 80.31 ± 9.27 | 82.03 ± 7.98 |
| Proposed: AGTCNet (0.0 s – 3.0 s Signal) | **66.82 ± 9.62** | 70.20 ± 9.51 | **82.88 ± 6.97** | 84.13 ± 6.56 |

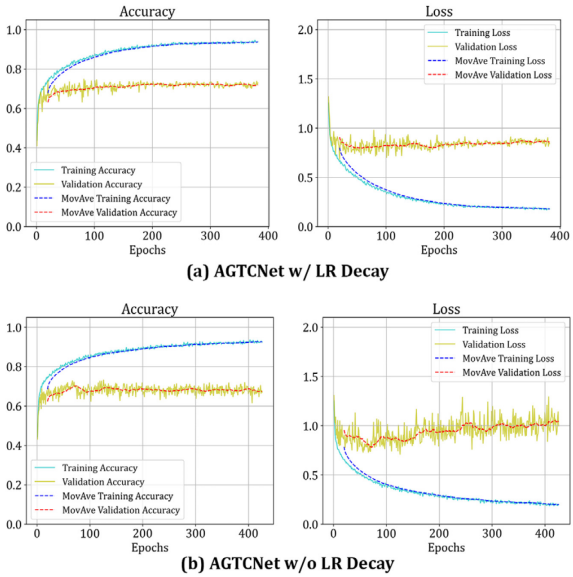

FIGURE 13. Learning curves of the baseline AGTCNet (-0.5 s – 4.0 s) with and without LR decay for subject-independent classification on the BCICIV2A dataset.

As for EEG signal preprocessing, the application of CAR resulted in a notable performance enhancement with a moving average accuracy increase of up to 2.62%. Similarly, downsampling the EEG signal from 250 Hz to 125 Hz led to an improved moving average accuracy of up to 1.99%. Both preprocessing techniques demonstrate clear benefits in enhancing the robustness of the EEG signal and improving the model's feature learning.

Moving on, the architecture of AGTCNet was systematically modified by removing or altering specific modules to assess their individual contributions to the model's overall performance. As detailed in the same table, the GCAT module significantly enhanced the model performance, achieving a remarkable boost in moving average accuracy of up to 3.7%, while the TCE module contributed an impressive improvement of up to 2.35%. In contrast, the removal of both modules resulted in a substantial performance drop, ranging from 2.84% to 4.17%. Notably, the GCAT, which leverages convolutional layers to jointly process the temporal dependencies of each EEG channel, outperformed GAT, which utilizes linear layers for message passing, with an improvement of up to 1.62%. Furthermore, incorporating a residual connection within the GCAT module enhanced the performance by up to 3.31%, demonstrating an effective balance between the aggregated channel information from adjacent channels and the original channel information. Aside from that, integrating a pooling layer into the CTC module improved the model performance by up to 2.44%, effectively reducing dimensionality and enhancing feature relevance.

Additionally, reducing the EEG signal duration from 4.5 seconds (-0.5 s – 4.0 s) to 3 seconds (0.0 s – 3.0 s) led to a 0.18% increase in the moving average accuracy for subject-independent classification and a 1.29% improvement for subject-specific classification. This enhancement is primarily attributed to the exclusion of irrelevant EEG signals prior to cue onset, which likely introduce noise, as well as the removal of weak MI stimuli, as subjects tend to lose concentration toward the end of the trials.

Overall, the ablation study highlights that the proposed AGTCNet architecture, in conjunction with the EEG signal preprocessing and model training strategy, significantly enhanced the robustness of MI-EEG classification. Notably, both the GCAT and TCE modules jointly enhance AGTCNet's ability to attentively and informatively capture expressive subject-invariant and session-invariant representations of EEG signals, as demonstrated by the substantial performance improvement attributed to the integration of these modules. Furthermore, the utilization of a 3-second EEG signal underscores the capability of AGTCNet to accurately distinguish MI tasks with short-duration EEG signals, in contrast to most existing models, which rely on longer signals for MI-EEG classification.

To further assess the effectiveness of AGTCNet in distinguishing MI tasks, t-distributed stochastic neighbor embedding (t-SNE) was used to visualize the learned latent representations prior to the final classification layer. First, kernel principal component analysis (KPCA) with a cosine kernel was applied to reduce the learned high-dimensional representations to 256 components. t-SNE was then employed to project these components into a two-dimensional space while preserving the local structure. Fig. 14 presents the t-SNE visualizations of AGTCNet for both subject-independent and subject-specific classifications. The resulting 2D projections revealed distinct clusters corresponding to different MI tasks with well-defined boundaries, demonstrating the model's ability to capture discriminative subject-invariant and session-invariant EEG representations, thereby effectively differentiating MI tasks using raw EEG signals.

## B. COMPARISON OF MODEL PERFORMANCE ON THE BCICIV2A DATASET

This section benchmarks the performance of the proposed AGTCNet against state-of-the-art models—EEGNet, EEG-TCNet, TCNet-Fusion, ATCNet, and DB-ATCNet—for both subject-independent and subject-specific classifications using the BCICIV2A dataset. In particular, each of these state-of-the-art models was reproduced using its publicly available code, following the preprocessing techniques and training methodology reported in the respective original paper. The performance results, specifically the moving average accuracy, accuracy, and kappa score, are summarized in Table 3. To evaluate the statistical significance of the performance improvements of AGTCNet, right-tailed Welch's t-tests were conducted against each of the reproduced state-of-the-art models, accounting for the differences in variance among the models. A detailed comparison of the model performance on individual subjects is provided in APPENDIX B.

The results show that AGTCNet outperformed all state-of-the-art models in both subject-independent and subject-specific classifications, achieving moving average accuracies of 66.82% and 82.88%, respectively, with lower variance. Notably, all p-values in subject-independent classification were below 0.05, indicating that the performance improvements of AGTCNet are statistically significant. Furthermore, AGTCNet demonstrated statistically superior performance in subject-specific classification compared to EEGNet, EEG-TCNet, and TCNet-Fusion, with p-values below 0.001.

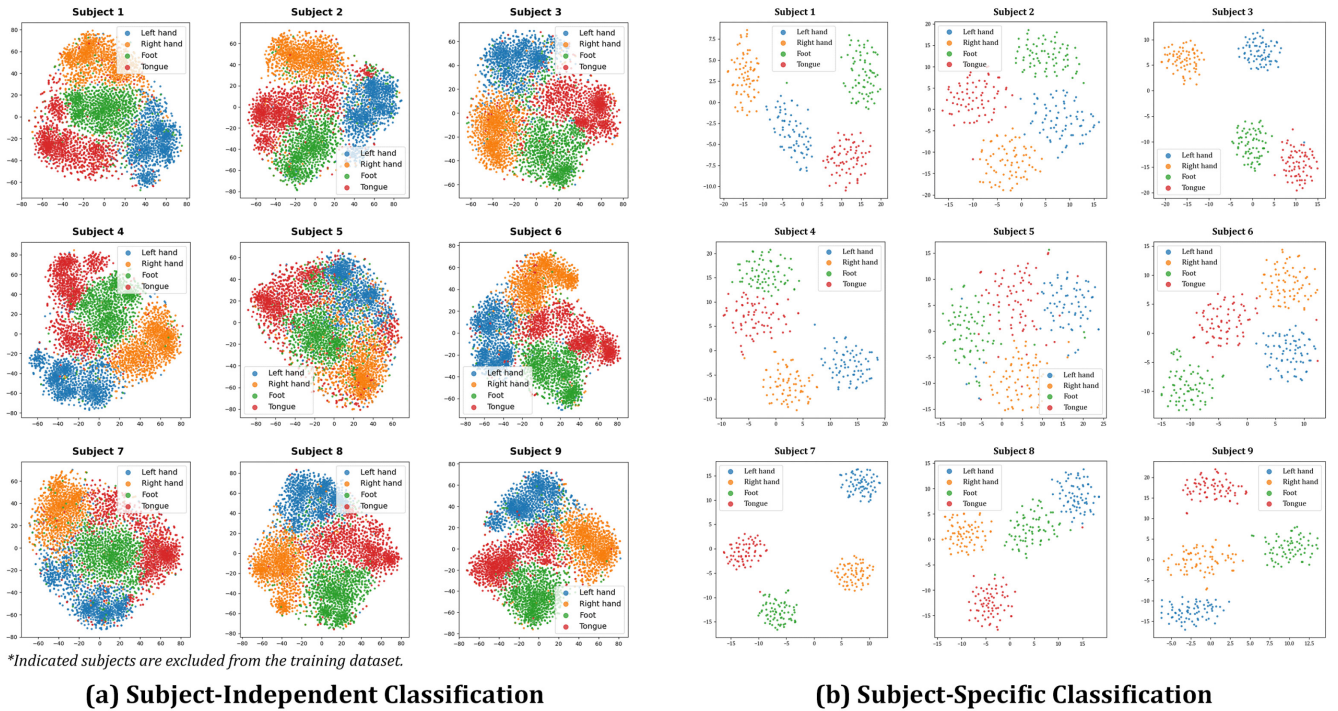

*Indicated subjects are excluded from the training dataset.

(a) Subject-Independent Classification    (b) Subject-Specific Classification

**FIGURE 14.** t-SNE visualizations of the learned latent representations from the baseline AGTCNet (-0.5 s – 4.0 s) for subject-independent and subject-specific classifications on the BCICIV2A dataset.

**TABLE 3.** Performance comparison of AGTCNet with reproduced state-of-the-art models for subject-independent and subject-specific classifications on the BCICIV2A dataset.

| Models | Subject-Independent | | | Subject-Specific | | |
|---|---|---|---|---|---|---|
| | MA Acc (%) | Acc (%) | κ-score | MA Acc (%) | Acc (%) | κ-score |
| EEGNet [40] | 61.77 ± 11.80** | 65.42 ± 11.60** | 0.54 ± 0.15** | 73.38 ± 11.04* | 75.50 ± 11.02* | 0.67 ± 0.15* |
| EEG-TCNet [45] | 62.80 ± 11.83** | 65.94 ± 11.59** | 0.55 ± 0.15** | 73.38 ± 11.62* | 75.87 ± 11.39* | 0.68 ± 0.15* |
| TCNet-Fusion [46] | 63.65 ± 11.24*** | 68.86 ± 10.93 | 0.58 ± 0.15 | 75.33 ± 9.62* | 77.81 ± 9.42* | 0.70 ± 0.13* |
| ATCNet [29] | 63.96 ± 10.74*** | 67.90 ± 10.32 | 0.57 ± 0.14 | 81.39 ± 8.30 | 83.40 ± 8.05 | 0.78 ± 0.11 |
| DB-ATCNet [30] | 62.67 ± 13.27** | 68.80 ± 12.32 | 0.58 ± 0.16 | 82.56 ± 7.90 | **84.80** ± 7.62 | **0.80** ± 0.10 |
| AGTCNet | **66.82 ± 9.62** | **70.20 ± 9.51** | **0.60 ± 0.13** | **82.88 ± 6.97** | 84.13 ± 6.56 | 0.79 ± 0.09 |
| AGTCNet (No LR Decay) | 66.42 ± 9.72 | 71.16 ± 9.26 | 0.62 ± 0.12 | 83.51 ± 6.39 | 85.06 ± 6.19 | 0.80 ± 0.08 |

Note: Asterisks indicate statistically significant improvements in AGTCNet over the corresponding state-of-the-art model, as determined by right-tailed Welch's t-test (*: $p < 0.001$, **: $p < 0.01$, ***: $p < 0.05$).

Although DB-ATCNet exhibited marginally higher accuracy and kappa score for subject-specific classification, the superior moving average accuracy of AGTCNet holds greater practical significance. As previously emphasized, both the standard accuracy and Cohen's kappa metrics are based only on the best-performing model checkpoint across all training epochs, leading to overestimation in the reported performance. In contrast, the moving average accuracy metric captures the underlying model's performance trend across training epochs, providing a more representative and reliable measure of the model's generalizability.

Notably, similar to other state-of-the-art models, both ATCNet and DB-ATCNet did not utilize LR decay, which allowed continued exploration of a wide parameter space and prevented the model from converging to a local optimum, resulting in considerable fluctuations in the validation learning curve, as depicted in Fig. 15. Conversely, AGTCNet employed LR decay, which progressively narrowed the parameter space, facilitating convergence to a local optimum. As illustrated in Fig. 16, this strategy promoted stable learning but with a slightly lower peak performance. Nevertheless, as noted in Table 3, AGTCNet outperformed DB-ATCNet across all three metrics when LR decay was not applied.

Fig. 17 compares the confusion matrices of AGTCNet with the reproduced state-of-the-art models for both subject-independent and subject-specific classifications. The confusion matrices reveal that AGTCNet exhibited greater intensity along the diagonal, indicating superior precision in classifying MI tasks.

In addition to classification performance, model efficiency is also a crucial consideration. Table 4 presents a comparison of the model complexity of AGTCNet with that of the state-of-the-art models, highlighting its efficiency. Specifically, AGTCNet utilized 49.87% fewer parameters than DB-ATCNet and 34.82% fewer than ATCNet. In terms of inference time, tests conducted on an Nvidia Quadro RTX 6000 GPU (24 GB memory) show that AGTCNet achieved 64.65% faster inference time than DB-ATCNet and 59.19% faster than ATCNet, as depicted in Fig. 18.

Along with its computational efficiency compared to ATCNet and DB-ATCNet, AGTCNet achieved statistically significant improvements in moving average accuracy of 4.15% ($p < 0.01$) and 0.32% for subject-independent and subject-specific classifications, respectively, as demonstrated in Fig. 19. This underscores the expressiveness of AGTCNet

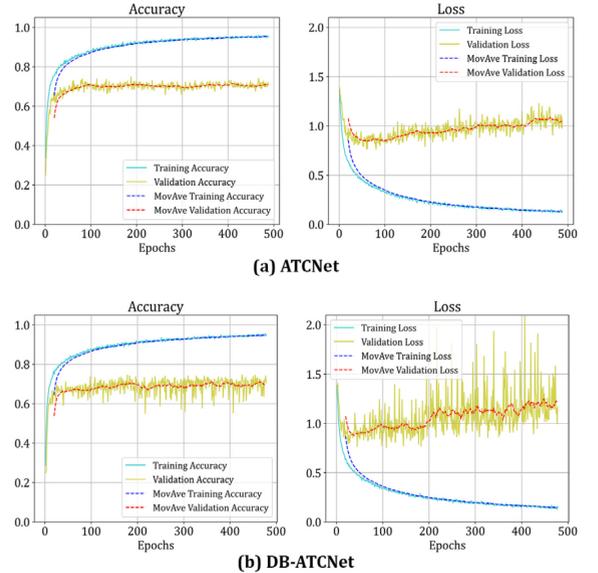

**FIGURE 15.** Learning curves of the reproduced ATCNet and DB-ATCNet for subject-independent classification on the BCICIV2A dataset.

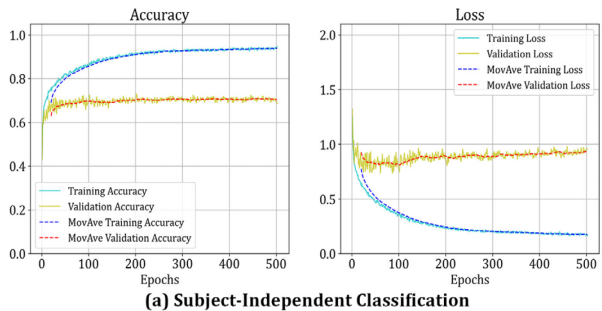

**FIGURE 16.** Learning curves of AGTCNet for subject-independent and subject-specific classifications on the BCICIV2A dataset.

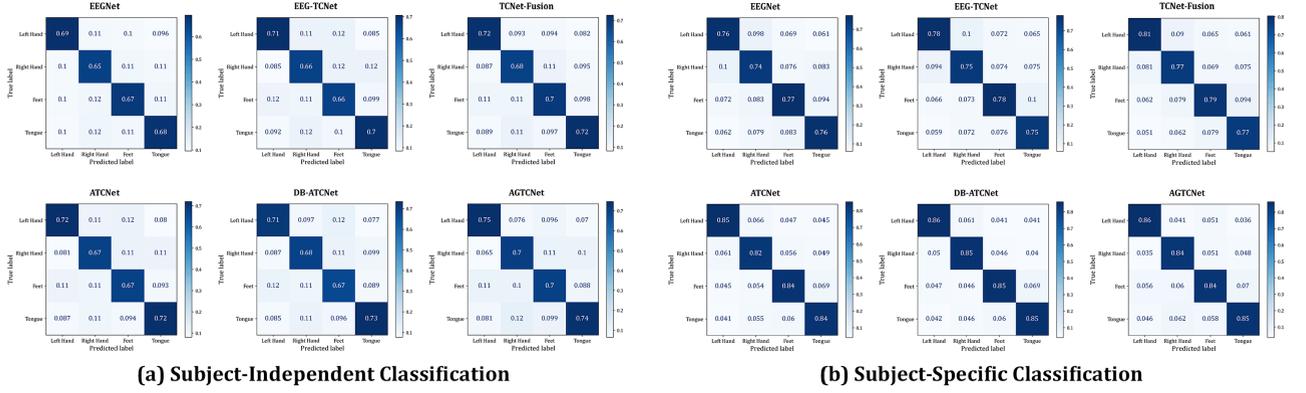

(a) Subject-Independent Classification    (b) Subject-Specific Classification

**FIGURE 17.** Comparison of the confusion matrices of AGTCNet with reproduced state-of-the-art models for subject-independent and subject-specific classifications on the BCICIV2A dataset.

**TABLE 4.** Comparison of the model complexity of AGTCNet with reproduced state-of-the-art models.

| Models | Parameters | Data Size (KB) | Inference Time (ms) |
|---|---|---|---|
| EEGNet [40] | 2,628 | 10.27 | 6.9957 ± 0.6941 |
| EEG-TCNet [45] | 4,272 | 16.69 | 20.4739 ± 1.1236 |
| TCNet-Fusion [46] | 17,584 | 68.69 | 21.2968 ± 1.3686 |
| ATCNet [29] | 115,172 | 449.89 | 95.2686 ± 2.5958 |
| DB-ATCNet [30] | 149,738 | 584.91 | 109.9737 ± 2.9906 |
| AGTCNet | 75,069 | 293.24 | 38.8768 ± 1.6511 |

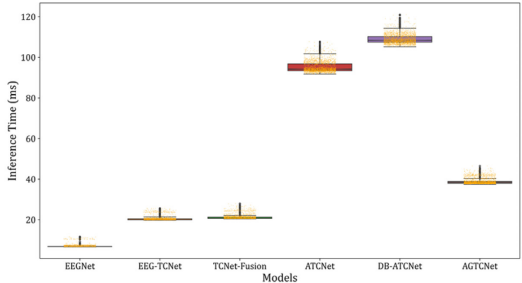

**FIGURE 18.** Comparison of the inference time of AGTCNet with reproduced state-of-the-art models.

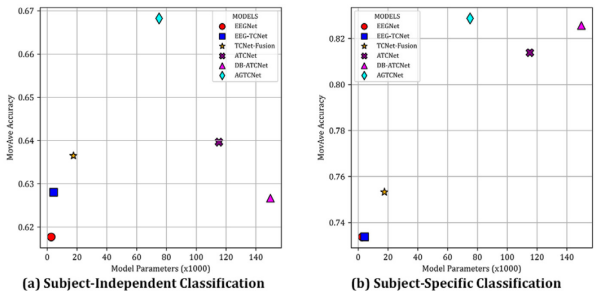

**FIGURE 19.** Scatterplot of moving average accuracy versus model parameters for AGTCNet and reproduced state-of-the-art models in subject-independent and subject-specific classifications on the BCICIV2A dataset.

in learning subject-invariant and session-invariant spatiotemporal EEG representations, addressing a key challenge in MI-EEG classification. As emphasized in [98], models benefit from prior knowledge to better understand the tasks they are designed to solve. However, most existing MI-EEG models, particularly those based solely on CNN and RNN architectures, fail to incorporate prior knowledge regarding the spatial relationships among EEG channels. AGTCNet addresses this gap by explicitly incorporating an inductive bias through the EEG channel adjacency graph, thereby enhancing the model's ability to distinguish EEG signals.

Furthermore, Table 5 presents a comprehensive comparison of the subject-independent and subject-specific classification performance of the proposed AGTCNet with models from recent studies benchmarked on the BCICIV2A dataset. The table also details the model architectures, input EEG signal timeframes, and preprocessing methods employed. Consistent with the performance metrics reported in their respective original papers, the comparative analysis is primarily conducted based on the average of the best accuracies and kappa scores achieved for each subject. The comparison demonstrates that AGTCNet outperformed all previous models in both subject-independent and subject-specific classifications, achieving state-of-the-art accuracies of 72.28% and 86.96%, with kappa scores of 0.63 and 0.83, respectively. Notably, AGTCNet surpassed DB-ATCNet by exhibiting more stable validation performance and lower variance.

### C. COMPARISON OF MODEL PERFORMANCE ON THE EEGMMIDB DATASET

This section further evaluates the performance of the proposed AGTCNet using the EEGMMIDB dataset, focusing on both the 4-class and 2-class MI-EEG classification tasks, as detailed in Table 6. The results show that AGTCNet achieved moving average accuracies of 64.14% and 85.22% for the 4-class and 2-class subject-independent classifications, respectively, as depicted in Fig. 20. Upon fine-tuning for subject-specific classification, the model demonstrated further performance improvement, with moving average accuracy reaching 72.13% for the 4-class classification task and 90.54% for the 2-class classification task.

Fig. 21 and 22 present the confusion matrices of AGTCNet for the 4-class and 2-class subject-independent and subject-specific classifications, respectively. Particularly, in the 4-class classification task, the confusion matrices reveal that AGTCNet effectively differentiated the "left fist," "right fist," and "both feet" tasks with high precision but struggles to distinguish the "both fists" task. This challenge is likely due to the overlapping dynamics of the "both fists" task with those of the "left fist" and "right fist" tasks, as well as its similarity to the "both feet" task, which involves simultaneous left and right actions, thus resulting in classification ambiguity. Meanwhile, AGTCNet achieved higher precision in the 2-class classification task, owing to its reduced number of classes, which simplifies the classification task.

**TABLE 5.** Performance comparison of AGTCNet with models from recent studies for subject-independent and subject-specific classifications on the BCICIV2A dataset.

| Model | Preprocessing [b] | Subject-Independent | | Subject-Specific | |
|---|---|---|---|---|---|
| | | Best Acc (%) | Best κ-score | Best Acc (%) | Best κ-score |
| **EEGNet**: CNN (2018) [40] [a] | -0.5 s – 4.0 s STD | 68.81 ± 11.86 | 0.58 ± 0.16 | 79.71 ± 10.67 | 0.73 ± 0.14 |
| **CRAM**: SWC, CNN, LSTM, Attention (2019) [47] | 0.0 s – 4.0 s | 59.10 ± 10.85 | — | — | — |
| **EEG-TCNet**: CNN, TCN (2020) [45] [a] | -0.5 s – 4.0 s STD | 69.64 ± 11.48 | 0.60 ± 0.15 | 80.83 ± 10.21 | 0.74 ± 0.14 |
| **NG-CRAM**: SWC, Defined Graph Topology, Symmetric Graph Filtering, CNN, LSTM, Attention (2020) [63] | 0.0 s – 4.0 s | 60.11 ± 9.96 | — | — | — |
| **TCNet-Fusion**: CNN, TCN (2021) [46] [a] | -0.5 s – 4.0 s STD | 71.12 ± 10.86 | 0.61 ± 0.14 | 80.94 ± 9.32 | 0.75 ± 0.12 |
| **CNN-BiLSTM**: Attention, Inception CNN, BiLSTM (2022) [28] | 8 Hz – 35 Hz BPF STD | — | — | 75.81 ± 13.86 | — |
| **ATCNet**: CNN, MHA, TCN (2023) [29] [a] | -0.5 s – 4.0 s STD | 70.68 ± 9.31 | 0.61 ± 0.12 | 86.46 ± 7.61 | 0.82 ± 0.10 |
| **CRAM + Baseline Correction Module (BCM) + Subject-Invariant Loss Function ($L_{invariant}$)** (2023) [51] | -0.5 s – 4.0 s | 61.82 ± 10.68 | — | — | — |
| **Global Adaptive Transformer**: CNN, MHA, GAN (2023) [33] | 0.0 s – 4.0 s 4 Hz – 40 Hz BPF STD | — | — | 76.58 ± 15.06 | 0.69 |
| **DB-ATCNet**: Dual-branch, CNN, MHA, TCN (2024) [30] [a] | -0.5 s – 4.0 s STD | 71.01 ± 11.94 | 0.61 ± 0.16 | **87.65 ± 6.96** | **0.84 ± 0.09** |
| **DMT-GAT**: Fully-Connected Graph Topology, CNN, GAT (2024) [64] | — | — | — | 82.20 ± 10.30 | — |
| **EEGAT**: KNN-based Graph Topology, CNN, GAT (2024) [34] | 4 Hz – 40 Hz BPF 128-Hz DS ExpMov STD | 51.72 ± 10.03 | — | — | — |
| **AGTCNet**: GCAT, CNN, MHA | 0.0 s – 3.0 s 125-Hz DS CAR | **72.28 ± 8.98** | **0.63 ± 0.12** | 86.96 ± 5.68 | 0.83 ± 0.08 |

[a] Reproduced model performance.
[b] BPF: Bandpass Filtering; DS: Downsampling; STD: Standardization.

**TABLE 6.** Performance of AGTCNet for the 4-class and 2-class subject-independent and subject-specific classifications on the EEGMMIDB dataset.

| Classes | Subject-Independent | | | Subject-Specific | | |
|---|---|---|---|---|---|---|
| | SMA Acc (%) | Acc (%) | κ-score | SMA Acc (%) | Acc (%) | κ-score |
| **4-Class** | 64.14 ± 2.83 | 65.44 ± 2.81 | 0.54 ± 0.04 | 72.13 ± 16.66 | 74.81 ± 15.70 | 0.66 ± 0.21 |
| **2-Class** | 85.22 ± 2.35 | 86.61 ± 2.20 | 0.73 ± 0.04 | 90.54 ± 10.49 | 92.53 ± 9.38 | 0.85 ± 0.19 |

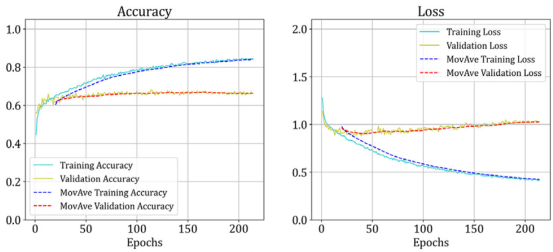

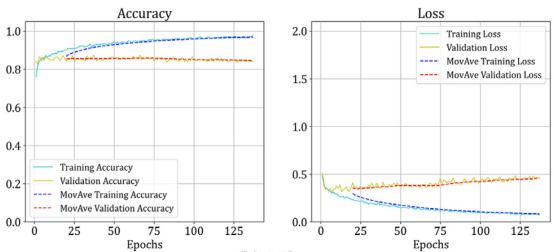

**FIGURE 20.** Learning curves of AGTCNet for the 4-class and 2-class subject-independent classifications on the EEGMMIDB dataset.

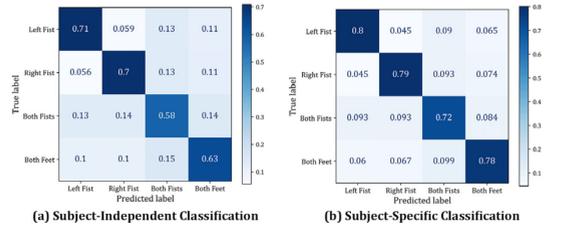

**FIGURE 21.** Confusion matrices of AGTCNet for the 4-class subject-independent and subject-specific classifications on the EEGMMIDB dataset.

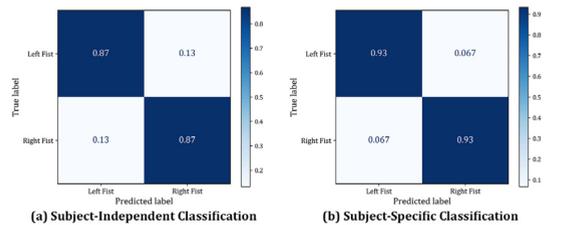

**FIGURE 22.** Confusion matrices of AGTCNet for the 2-class subject-independent and subject-specific classifications on the EEGMMIDB dataset.

In comparison to models from recent studies benchmarked on the EEGMMIDB dataset, Table 7 presents a comprehensive comparison of the subject-independent and subject-specific classification performance, based on the average of the best accuracies achieved across each fold, as reported in their respective original papers. The table also includes detailed information on the model architectures, input EEG signal timeframes, preprocessing methods employed, as well as the number of subjects and classes considered. As noted in the table, previous studies have focused on varying sets of MI tasks, with some using the resting state derived from baseline recordings rather than the "both fists" task, likely due to overlapping dynamics associated with the latter [37], [50], [54]. Moreover, these studies evaluated the model performance using varying numbers of subjects. To ensure a fair comparison, the comparative analysis was conducted separately for the 4-class and 2-class classification tasks. The comparison demonstrates that AGTCNet outperformed all previous models in both the 4-class and 2-class subject-independent classifications, achieving state-of-the-art accuracies of 66.20% and 87.12%, respectively. Furthermore, AGTCNet also achieved superior performance in subject-specific classification, surpassing existing models that employed fine-tuning, with accuracies of 81.37% and 96.99% for the 4-class and 2-class classification tasks, respectively.

## VII. CONCLUSION

This study introduces the attentive graph-temporal convolutional network (AGTCNet), a novel graph-temporal architecture for subject-independent and subject-specific MI-EEG classifications. By incorporating inductive bias through an EEG channel adjacency graph based on the anteroposterior and mediolateral topographic configuration of EEG electrodes, AGTCNet effectively learns expressive subject-invariant and session-invariant spatiotemporal EEG representations. Specifically, the proposed graph-convolutional attention network (GCAT) facilitates informative message passing among adjacent EEG channels, jointly capturing the spatial and temporal dynamics of multichannel EEG signals. Furthermore, the graph-convolutional attention pooling (GCAP) adaptively aggregates the multichannel EEG representations into monochannel representations, capturing the relative relevance of each individual EEG channel across multiple subspaces.

The ablation study highlights the contributions of the proposed AGTCNet architecture, in conjunction with the EEG signal preprocessing and model training strategy. Notably, the proposed AGTCNet significantly outperformed existing MI-EEG classification models on different datasets acquired from varying EEG headsets, achieving state-of-the-art performance while utilizing a compact architecture. In particular, with a 49.87% reduction in model size, 64.65% faster inference time, and shorter input EEG signal, AGTCNet achieved a moving average accuracy of 66.82% ($p < 0.01$) for subject-independent classification on the BCICIV2A dataset. Moreover, fine-tuning for subject-specific classification further enhanced the model performance, with the moving average accuracy reaching 82.88%. Meanwhile, on the EEGMMIDB dataset, AGTCNet achieved moving average accuracies of 64.14% and 85.22% for the 4-class and 2-class subject-independent classifications, respectively, with further improvements to 72.13% and 90.54% for subject-specific classifications. The robust performance and computational efficiency of AGTCNet highlight its effectiveness in distinguishing MI tasks from raw EEG signals, underscoring its practicality for real-time deployment in resource-constrained BCI systems.

**TABLE 7.** Performance comparison of AGTCNet with models from recent studies for the 4-class and 2-class subject-independent and subject-specific classifications on the EEGMMIDB dataset.

| Model | Preprocessing [a] | Subjects | Class [b] | Best Acc (%) Subject-Independent | Best Acc (%) Subject-Specific |
|---|---|---|---|---|---|
| **CNNs** (2018) [37] | 0.0 s – 3.0 s | 105 | 4 (L/R/0/F) | 58.59 ± 14.67 | 68.51 ± 12.56 |
| **EEGNet**: CNN (2020) [54] | 0.0 s – 3.0 s | 105 | 4 (L/R/0/F) | 65.07 | 70.83 |
| **ConTraNet**: CNN, Transformer (2022) [50] | — | 105 | 4 (L/R/0/F) | 65.44 ± 2.76 | — |
| **ST-GNN**: Trainable Graph Topology, CNN, GCN (2022) [31] | 0.0 s – 3.0 s CAR + STD | 109 | 4 | 64.38 | — |
| **SGLNet**: KNN-based Graph Topology, Spike Encoder, SNN, GCN, LSTM (2023) [32] | 0.0 s – 4.0 s CAR + STD | 108 | 4 | 57.72 ± 3.42 (10-Fold) | — |
| **AGTCNet**: GCAT, CNN, MHA | 0.0 s – 4.0 s CAR | 103 | 4 | **66.20 ± 2.95** | **81.37 ± 13.03** |
| **CNNs** (2018) [37] | 0.0 s – 3.0 s | 105 | 2 | 80.38 ± 12.54 | 86.49 ± 10.09 |
| **EEGNet**: CNN (2020) [54] | 0.0 s – 3.0 s | 105 | 2 | 82.43 | 84.32 |
| **DG-CRAM**: SWC, Defined Graph Topology, Symmetric Graph Filtering, CNN, LSTM, Attention (2020) [63] | 0.0 s – 3.1 s | 105 | 2 | 74.71 ± 4.19 (9-Fold) | — |
| **ConTraNet**: CNN, Transformer (2022) [50] | — | 105 | 2 | 83.61 ± 2.38 | — |
| **ST-GNN**: Trainable Graph Topology, CNN, GCN (2022) [31] | CAR + STD | 109 | 2 | 82.92 | — |
| **AGTCNet**: GCAT, CNN, MHA | 0.0 s – 4.0 s CAR | 103 | 2 | **87.12 ± 1.98** | **96.99 ± 5.19** |

[a] STD: Standardization.
[b] 'L': Left Hand; 'R': Right Hand; '0': Rest State; 'F': Both Feet.

Overall, this study demonstrates the efficacy of integrating GCAT with an EEG channel adjacency graph that models the spatial relationships among EEG channels to jointly learn intricate spatiotemporal EEG representations, thereby enhancing the expressiveness of the model. Future research may build upon this work by exploring the integration of RNNs with GAT [99]. It is highly recommended for future work to employ learning rate decay to facilitate stable learning, mitigate overfitting, and yield a robust model. Finally, to ensure representative and generalized performance evaluations as well as comparability across studies, it is imperative that future studies adhere to the established practical model training-evaluation frameworks and utilize the moving average accuracy metric. In particular, the SN training-evaluation framework should be employed for the development of subject-independent EEG classifiers, whereas for subject-specific EEG classifiers, the SL-DS or SL-DS-FT training-evaluation framework should be utilized, with the latter leveraging fine-tuning to minimize reliance on large subject-specific datasets.

## APPENDIX A
## AGTCNET PARAMETER CONFIGURATIONS

Table 8 details the parameter configurations of the proposed AGTCNet, which were systematically optimized through an extensive hyperparameter tuning procedure. The hyperparameter search space included the layer type (i.e., linear, standard convolution, and depthwise separable convolution), number of layers, layer parameters—specifically filters, kernel size, strides, pool size, and units—, activation function (i.e., ReLU, ELU, SELU, Leaky ReLU, and PReLU), normalization technique (i.e., batch normalization and layer normalization), dropout rate, loss function (i.e., categorical cross-entropy and LogE), optimizer (i.e., Adam, AdamW, AdaMax, and NAdam), learning rate, batch size, as well as the LR decay factor and patience.

Specifically, the maximum moving average validation accuracy, as established in Section IV.D, was used as the principal performance metric during model hyperparameter tuning to ensure stable and consistent performance, rather than being attributed to the stochastic nature of model training. In accordance with the primary objective of BCI research, the proposed AGTCNet was tuned for the 4-class subject-independent MI-EEG classification task using the BCICIV2A dataset, with subject 1 as the validation dataset. Subsequently, the tuned model was cross-validated on the remaining subjects (i.e., subjects 2 to 9) and further evaluated across various classification tasks and on a different dataset (i.e., EEGMMIDB) to assess its generalizability and robustness.

Notably, among the activation functions, the Scaled Exponential Linear Unit (SELU) consistently demonstrated superior performance during hyperparameter tuning. Although the Rectified Linear Unit (ReLU) is a widely employed activation function in deep learning to mitigate the vanishing gradient problem, it suffers from the "dying ReLU" issue, wherein neurons receiving negative inputs produce outputs of zero, rendering them inactive and causing a significant portion of the network to cease learning [100]. On the other hand, similar to the Exponential Linear Unit (ELU), Leaky ReLU, and Parametric ReLU (PReLU), SELU addresses this limitation by introducing a small negative gradient $\alpha$ for negative inputs, ensuring that all neurons remain active. Moreover, the self-normalizing properties of SELU facilitate the stabilization of the model's learning process and accelerate convergence, rendering it particularly effective in deep neural networks. These properties thus collectively establish SELU as the optimal activation function for the proposed AGTCNet.

Additionally, various EEG channel adjacency graph topologies were also explored during model tuning, specifically predefined, fully-connected, and trainable graphs. Among these, the predefined graph topology demonstrated superior performance. In contrast, while the trainable graph topology offers a promising approach for capturing complex relationships among EEG channels, it failed to converge. This can be attributed to the exponential increase in parameter space, which significantly complicates the model's learning process, especially under the constraints of a limited dataset size.

## APPENDIX B
## COMPARISON OF INDIVIDUAL SUBJECT MODEL PERFORMANCE ON THE BCICIV2A DATASET

Tables 9 and 10, respectively, present comprehensive evaluations of the subject-independent and subject-specific classification performance of the proposed AGTCNet against reproduced state-of-the-art models on individual subjects from the BCICIV2A dataset, based on the mean moving average accuracy across 10 training trials. The results demonstrate that AGTCNet statistically outperformed the state-of-the-art models in both classification tasks across the majority of subjects, with AGTCNet surpassing DB-ATCNet in most subjects. This robust performance underscores the generalizability of AGTCNet in classifying MI-EEG signals across diverse users and time periods.

Moreover, Fig. 23 also presents the confusion matrices of AGTCNet for both subject-independent and subject-specific classifications on individual subjects. Notably, the precision of MI-EEG classification varies across subjects, which can be attributed to the nuances in individual neural dynamics and BCI literacy levels [101]. In general, the subject-specific MI-EEG classifiers exhibited greater intensity along the diagonal, indicating enhanced performance when tuned to individual subjects, as compared to the uncalibrated subject-independent MI-EEG classifiers.

**TABLE 8.** AGTCNet module parameter configurations.

| Module Components | | Parameters |
|---|---|---|
| **Channel-wise Temporal Convolution (CTC) Module** | | |
| | ctconv | Conv(filters=8, kernel=(1,32), strides=(1,1), padding='valid', bias=False) |
| | | BatchNorm |
| | ctpool | AvePool(pool_size=(1,4), strides=(1,2)) |
| **Graph Convolutional Attention Network (GCAT) Module** | | |
| | mod_weight * | DSConv(filters=16, kernel=(1,8), depth=4, strides=(1,1), padding='same', bias=False) |
| | mod_val * | BatchNorm |
| | | SELU |
| | mod_attn_src * | DSConv(filters=1, kernel=(1,2), depth=4, strides=(1,1), padding='same', bias=False) |
| | mod_attn_dst * | BatchNorm |
| | | SELU |
| * Unique instance for each head. | | |
| | gcat | GCAT(filters=16, heads=2, concat_heads=False, self_loops=True, attn_actvn=PReLU(shared_axes=[1]), attn_dropout_rate=0.2, bias=True) |
| | | BatchNorm |
| | | SELU |
| | | Dropout(0.25) |
| **Global Convolutional Adaptive Pooling (GCAP) Module** | | |
| | chpool | DWConv(kernel=(NUM_CHANNELS, 1), depth=2, strides=(1,1), padding='valid', bias=False, depthwise_constraint=MaxNorm(1., axis=0)) |
| | | BatchNorm |
| | | SELU |
| **Global Temporal Convolution (GTC) Module** | | |
| | tpool1 | AvePool(pool_size=(1,4), strides=(1,4)) |
| | | Dropout(0.25) |
| | tconv | DSConv(filters=96, kernel=(1,8), depth=4, strides=(1,1), padding='same', bias=False) |
| | | BatchNorm |
| | | SELU |
| | tpool2 | AvePool(pool_size=(1,4), strides=(1,4)) |
| | | Dropout(0.25) |
| **Temporal Context Enhancement (TCE) Module** | | |
| | tce_pos_encoder | PositionalEncoding(trainable_scale=True, scale_initializer='Zeros', scale_constraint=MinMaxValue(min=0., max=1.)) |
| | tce_mha | MHA(heads=2, key_dim=8, value_dim=8, dropout=0.6, bias=True, attention_axes=None) |
| | | BatchNorm |
| | | Dropout(0.3) |
| | tce_conv | DSConv(filters=96, kernel=(1,2), depth=4, strides=(1,1), padding='same', bias=False) |
| | | BatchNorm |
| | | SELU |
| | | Dropout(0.2) |
| **Classification Module** | | |
| | classifier | Linear(units=NUM_CLASSES, bias=True, kernel_constraint=MaxNorm(0.25, axis=0)) |
| | | Softmax |

**TABLE 9.** Comparison of the individual subject moving average accuracy of AGTCNet with reproduced state-of-the-art models for subject-independent classification on the BCICIV2A dataset.

| Models \ Subjects | EEGNet [40] | EEG-TCNet [45] | TCNet-Fusion [46] | ATCNet [29] | DB-ATCNet [30] | AGTCNet |
|---|---|---|---|---|---|---|
| 1 | 66.41 ± 2.30* | 68.99 ± 1.44** | 66.86 ± 1.16* | 69.17 ± 1.33*** | **70.91 ± 1.26** | 70.40 ± 0.32 |
| 2 | 45.49 ± 0.91* | 47.33 ± 1.83* | 47.29 ± 1.31* | 49.00 ± 1.71* | 44.36 ± 1.48* | **55.32 ± 1.36** |
| 3 | 78.92 ± 2.19* | 80.55 ± 2.53*** | 78.74 ± 0.69* | 78.56 ± 1.26* | 79.51 ± 1.20* | **82.70 ± 1.72** |
| 4 | 56.26 ± 1.57 | 56.86 ± 2.57 | **57.17 ± 1.54** | 56.78 ± 2.08 | 53.60 ± 1.60** | 55.99 ± 1.71 |
| 5 | 48.75 ± 2.82* | 52.64 ± 3.15* | 52.10 ± 2.83* | 58.23 ± 2.74*** | 54.62 ± 0.95* | **60.33 ± 1.65** |
| 6 | 49.05 ± 1.52* | 47.40 ± 1.81* | 51.66 ± 1.26* | 48.09 ± 1.90* | 42.77 ± 2.27* | **56.17 ± 2.06** |
| 7 | 70.30 ± 1.50* | 70.51 ± 2.07* | 71.35 ± 0.76* | 71.36 ± 1.28* | 70.66 ± 1.25* | **73.91 ± 0.90** |
| 8 | 75.41 ± 2.22 | 76.09 ± 1.98 | 77.26 ± 1.46 | 75.33 ± 1.29 | **77.37 ± 1.24** | 75.45 ± 1.39 |
| 9 | 65.36 ± 2.86* | 64.86 ± 3.33* | 70.43 ± 1.56 | 69.14 ± 1.28** | 70.19 ± 1.42 | **71.12 ± 1.13** |
| **OVERALL** | 61.77 ± 11.80** | 62.80 ± 11.83** | 63.65 ± 11.24*** | 63.96 ± 10.74*** | 62.67 ± 13.27** | **66.82 ± 9.62** |

Note: Asterisks indicate statistically significant improvements in AGTCNet over the corresponding state-of-the-art model, as determined by right-tailed Welch's t-test (*: $p < 0.001$, **: $p < 0.01$, ***: $p < 0.05$).

**TABLE 10.** Comparison of the individual subject moving average accuracy of AGTCNet with reproduced state-of-the-art models for subject-specific classification on the BCICIV2A dataset.

| Models<br>Subjects | EEGNet [40] | EEG-TCNet [45] | TCNet-Fusion [46] | ATCNet [29] | DB-ATCNet [30] | AGTCNet |
|---|---|---|---|---|---|---|
| 1 | 83.30 ± 1.59* | 80.92 ± 1.96* | 81.45 ± 1.77* | 85.07 ± 1.69*** | 85.25 ± 1.51*** | **86.75 ± 1.69** |
| 2 | 59.44 ± 3.24* | 56.89 ± 6.34* | 61.56 ± 2.41* | 67.63 ± 2.32* | 68.37 ± 1.67* | **72.50 ± 1.66** |
| 3 | 86.06 ± 4.35* | 87.08 ± 3.54* | 86.02 ± 2.82* | 91.47 ± 2.35*** | 91.79 ± 2.28 | **93.15 ± 1.59** |
| 4 | 62.10 ± 2.93* | 66.25 ± 2.16* | 66.18 ± 1.86* | 78.57 ± 1.80 | **81.45 ± 2.44** | 78.88 ± 1.32 |
| 5 | 68.79 ± 2.20* | 68.41 ± 3.91* | 72.70 ± 1.65*** | 78.19 ± 1.83 | **79.17 ± 1.82** | 76.50 ± 5.11 |
| 6 | 58.41 ± 1.84* | 57.14 ± 2.36* | 62.38 ± 1.72* | 70.26 ± 2.11* | 73.13 ± 1.73** | **76.58 ± 3.01** |
| 7 | 85.48 ± 2.18 | 86.54 ± 3.93 | 87.64 ± 1.67 | 91.69 ± 2.14 | **93.05 ± 2.13** | 86.44 ± 1.72 |
| 8 | 79.39 ± 3.30* | 79.98 ± 2.71* | 80.52 ± 1.89* | 85.40 ± 0.96* | 85.78 ± 1.03* | **88.47 ± 0.99** |
| 9 | 77.40 ± 2.78* | 77.24 ± 4.75* | 79.48 ± 2.31* | 84.20 ± 3.54 | 85.08 ± 3.30 | **86.61 ± 2.32** |
| OVERALL | 73.38 ± 11.04* | 73.38 ± 11.62* | 75.33 ± 9.62* | 81.39 ± 8.30 | 82.56 ± 7.90 | **82.88 ± 6.97** |

Note: Asterisks indicate statistically significant improvements in AGTCNet over the corresponding state-of-the-art model, as determined by right-tailed Welch's t-test (*: $p < 0.001$, **: $p < 0.01$, ***: $p < 0.05$).

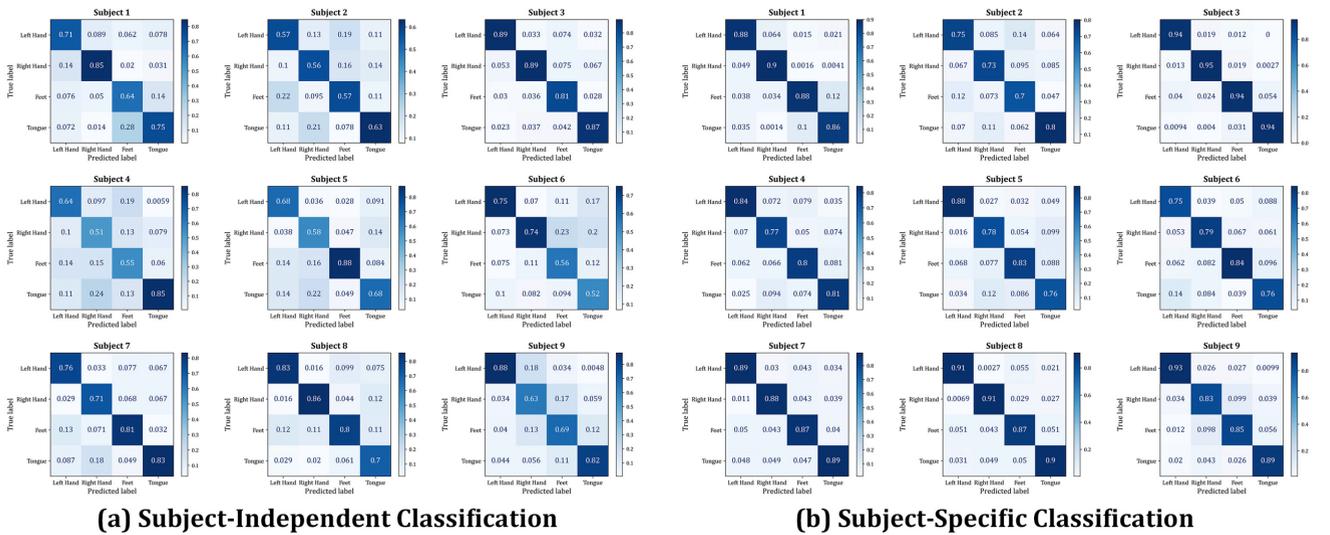

(a) Subject-Independent Classification   (b) Subject-Specific Classification

**FIGURE 23.** Individual subject confusion matrices of AGTCNet for subject-independent and subject-specific classifications on the BCICIV2A dataset.

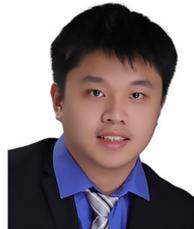

**GALVIN BRICE S. LIM** (Student Member, IEEE) obtained his M.S. and B.S. in Electronics and Communications Engineering from the De La Salle University – Manila, where he was awarded the Gold Medal for Outstanding Graduate Thesis. His research interests include brain-computer interface, deep learning, machine learning, robotics, electronic systems design, embedded systems, and the Internet of Things.

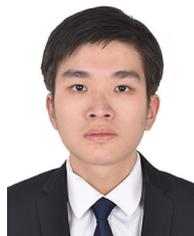

**BRIAN GODWIN S. LIM** is currently pursuing his Ph.D. in Information Science at the Nara Institute of Science and Technology. He obtained his Master and B.S. in Applied Mathematics, Major in Mathematical Finance from the Ateneo de Manila University. His research interests include mathematical modeling, graph theory, discrete algorithms, mathematical finance, econometrics, and machine learning.

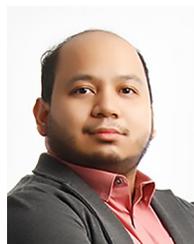

**ARGEL A. BANDALA** (Member, IEEE) is a full professor and research faculty member of the Department of Electronics and Computer Engineering at the De La Salle University – Manila. He obtained his Ph.D. and M.S. in Electronics and Communications Engineering from the same institution. In 2017, he received the 3rd prize in the Talent Search for Young Scientists of the National Academy of Science and Technology (NAST). In 2020, he was also one of the recipients of the NAST Outstanding Young Scientist (OYS) award. His research interests include artificial intelligence, swarm robotics, swarm intelligence, and meta-heuristics algorithms.

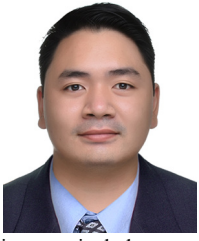

**JOHN ANTHONY C. JOSE** (Member, IEEE) is an associate professor of the Department of Electronics and Computer Engineering at the De La Salle University – Manila. He obtained his Ph.D., M.S., and B.S. in Electronics and Communications Engineering from the same institution. In 2014, he received the IEEE Outstanding Student Award in Region 10 Conference. In 2015, he also received the Outstanding Master's Thesis Award. His research interests include computer vision and deep learning.

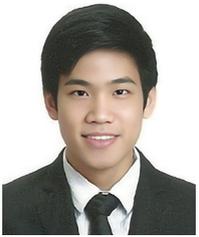

**TIMOTHY SCOTT C. CHU** is a faculty member and researcher of the Department of Mechanical Engineering at the De La Salle University – Manila, specializing in Mechanical and Mechatronics Engineering. He obtained his M.Sc. in Robotics Engineering from the Liverpool Hope University, where his research focused on brain-computer interface (BCI) drone operations. He also holds an M.S. in Mechanical Engineering from the De La Salle University – Manila, where he advanced his research on unmanned aerial vehicles (UAVs). His research interests include artificial intelligence, robotics, human-robot interaction, multi-UAV systems, and service robots, with a focus on improving human-robot collaboration and addressing pressing societal challenges.

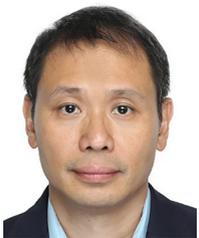

**EDWIN SYBINGCO** (Senior Member, IEEE) is a full professor of the Department of Electronics and Computer Engineering at the De La Salle University – Manila. He obtained his Ph.D., M.S., and B.S. in Electronics and Communications Engineering from the same institution. In 1996, he was trained in signal processing during his post-M.S. at the University of New South Wales, sponsored by the Department of Science and Technology (DOST). In 1997, he was an exchange scientist at the Tokyo Institute of Technology, funded by the Japan International Cooperation Agency (JICA). In 2016 and 2017, he was also an exchange scientist at the University of Arizona, sponsored by the United States Agency for International Development (USAID) under the Science, Technology, Research, and Innovation for Development (STRIDE) CARWIN Grant, for the projects entitled "Automation of Coco-Sugar Production" and "Smart Farm for All-Seasoned Tomato Production," respectively. Currently, he is involved in several government projects sponsored by the DOST Philippine Council for Industry, Energy and Emerging Technology Research and Development (PCIEERD). His research interests include computational intelligence, machine vision, robotics, and signal processing.